\documentclass{article}

\usepackage{tikz}
\usepackage{pgfplots}
\usepackage{PRIMEarxiv}
\usepackage{amsmath,amssymb,amsfonts}
\usepackage[utf8]{inputenc} 
\usepackage[T1]{fontenc}    
\usepackage{hyperref}       
\usepackage{url}            
\usepackage{booktabs}       
\usepackage{amsfonts}       
\usepackage{nicefrac}       
\usepackage{subcaption}
\usepackage{multirow}
\usepackage{color,soul}
\usepackage{algorithm}
\usepackage{algpseudocode}
\usepackage{xspace}
\usepackage{bm}
\usepackage{microtype}      
\usepackage{lipsum}
\usepackage{fancyhdr}       
\usepackage{graphicx}       
\graphicspath{{media/}}     
\usepackage{float}

\title{Deep Muscle EMG construction using A Physics-Integrated Deep Learning approach
}

\author{
  Rajnish Kumar\\
  Department of Applied Mechanics \\
  Indian Institute of Technology \\
  New Delhi\\
  \texttt{Rajnish.Kumar@am.iitd.ac.in}\\
  \And
  Tapas Tripura\\
  Department of Applied Mechanics \\
  Indian Institute of Technology \\
  New Delhi\\
  \texttt{tapas.t@am.iitd.ac.in}\\
   \And
  Souvik Chakraborty \\
  Department of Applied Mechanics \\
  Indian Institute of Technology\\
  New Delhi\\
  \texttt{souvik@am.iitd.ac.in} \\
   \And
  Sitikantha Roy \\
  Department of Applied Mechanics \\
  Indian Institute of Technology\\
  New Delhi\\
  \texttt{sroy@am.iitd.ac.in}
}

\begin{document}
\maketitle

\begin{abstract}
Electromyography (EMG)--based computational musculoskeletal modeling is a non-invasive method for studying musculotendon function, human movement, and neuromuscular control, providing estimates of internal variables like muscle forces and joint torques. However, EMG signals from deeper muscles are often challenging to measure by placing the surface EMG electrodes and unfeasible to measure directly using invasive methods. The restriction to the access of EMG data from deeper muscles poses a considerable obstacle to the broad adoption of EMG-driven modeling techniques. 
A strategic alternative is to use an estimation algorithm to approximate the missing EMG signals from deeper muscle. A similar strategy is used in physics-informed deep learning, where the features of physical systems are learned without labeled data. 
In this work, we propose a hybrid deep learning algorithm, namely the neural musculoskeletal model (NMM), that integrates physics-informed and data-driven deep learning to approximate the EMG signals from the deeper muscles. While data-driven modeling is used to predict the missing EMG signals, physics-based modeling engraves the subject-specific information into the predictions. 
Experimental verifications on five test subjects are carried out to investigate the performance of the proposed hybrid framework. The proposed NMM is validated against the joint torque computed from 'OpenSim' software. The predicted deep EMG signals are also compared against the state-of-the-art muscle synergy extrapolation (MSE) approach, where the proposed NMM completely outperforms the existing MSE framework by a significant margin.
\end{abstract}


\section{Introduction}
The electromyography (EMG)-based computational musculoskeletal (MSK) modeling is a non-invasive technique for studying musculotendon function, human movement, and neuromuscular control. It provides an approach to estimate internal variables, including muscle forces and joint torques, which are essential for various applications such as rehabilitation, sports performance, clinical diagnostics, and human-machine interaction\cite{zajac1989muscle,lloyd2003emg,buchanan2004neuromusculoskeletal,pizzolato2015ceinms,shourijeh2020advances,durandau2020towards}. 
Integrating MSK modeling with real-time physiological variables, such as personalized muscle activation data, can significantly enhance the applicability and adoption of these simulations. Surface electromyography (sEMG) signals are commonly used as an indirect method to estimate muscle activation patterns in the muscles of interest \cite{lacquaniti2012patterned,cappellini2020clinical,sartori2014hybrid,zonnino2019model}. 
However, the inability of sEMG to record an entire set of EMG signals from each of the relevant muscles is a major limitation \cite{ao2020evaluation,tahmid2024upper}.
This limitation arises due to challenges in recording deep muscle EMG signals, missing EMG data caused by measurement constraints, or an insufficient number of EMG channels when using sparse sensors. 
Although fine wire electrodes can detect deep muscle activity, they are invasive and can cause user discomfort \cite{ao2024comparison}. The inability to record signals from all relevant muscles can lead to an overestimation of individual muscle forces, ultimately reducing the overall accuracy of the MSK model \cite{sartori2014hybrid,ao2020evaluation,wang2023computational}.

In the literature, two primary strategies are employed to estimate muscle activations or forces, including those from deep or missing muscles: high-dimensional and low-dimensional approaches. 
The high-dimensional strategy involves model-based techniques, such as inverse dynamics and forward dynamic simulations. Inverse dynamics methods, such as static optimization (SO) \cite{kian2019static, anderson2001static,heintz2007static,seth2018opensim} and computed muscle control (CMC) \cite{thelen2003generating,delp2007opensim, seth2018opensim}, solve redundancy problems using optimization approaches in high-dimensional spaces to estimate muscle activation patterns that closely align with the desired kinetics or kinematics.  
Forward dynamic simulations, on the other hand, utilize EMG-driven MSK models to estimate muscle activations. 
However, EMG signals from deep or missing muscles often restrict its application \cite{sartori2012emg,zonnino2019model,ao2022emg}. 
To overcome this issue, deep muscle activations are estimated using hybrid techniques that combine static optimization with EMG-driven MSK modeling \cite{sartori2014hybrid,pizzolato2015ceinms}. 
Nevertheless, these methods increase computational and dimensional complexity due to the need to solve intricate MSK models and equations of motion for multiple body segments while accounting for muscle redundancy. 
Moreover, high-dimensional approaches face a large challenge in selecting biologically consistent objective functions \cite{pandy1995optimal,erdemir2007model}. 
The complexity of human movement dynamics and our incomplete knowledge of how the central nervous system organizes and performs motor tasks are the causes of this difficulty \cite{kim2009evaluation,rane2019deep}.

An alternative method to estimate EMG data from deep or missing muscles is the low-dimensional approach known as muscle synergy extrapolation (MSE) \cite{bianco2018can,rabbi2020non}. In MSE the experimentally recorded sEMG data are dimensionally reduced down into a smaller set of time-varying synergy excitation vectors and time-invariant weight vectors. Synergy excitation vectors represent the basis functions, reflecting the patterns used by the central nervous system (CNS) to generate movement. Subsequently, these basis functions are used to reconstruct the EMG data of missing muscles\cite{ao2020evaluation, ao2022emg,rabbi2024muscle,tahmid2024upper}.
Despite its utility, MSE faces several challenges. First, the effectiveness of synergy analysis is highly dependent on methodological choices, such as the selection and processing of EMG data, the number of synergies \cite{steele2013number}, and the matrix decomposition techniques employed \cite{turpin2021improve,ao2020evaluation}. Second, deriving basis functions from an incomplete set of EMG data does not guarantee the correctness of the resulting synergies \cite{van2023handling}. The accuracy of the basis vectors depends on the proportion and pattern of missing data \cite{rubin1976inference,little2019statistical}.  Methods like expectation-maximization principal component analysis (EM-PCA) \cite{ghahramani1995learning}, matrix factorization with missing values, or weighted PCA can be designed to extract valid basis vectors from observed data without imputing missing values. These methods can generate accurate basis vectors if the missing data is random (MAR) or adequately handled. However, if the missing data follow a more complex pattern, such as whole EMG signal from a deep muscle being "missing not at random" (MNAR), the basis vectors may become biased and fail to fully capture the underlying data structure \cite{santos2019generating}.

A pragmatic solution to these challenges could be to construct EMG signals directly from joint kinematics data. For example, an RNN-based model has been proposed to predict upper limb muscle activity using motion parameters such as joint angle and velocity \cite{schmidt2023concepts}. More recently, a hybrid generative model has been developed to synthesize intramuscular electromyography signals from kinematic data, effectively capturing the spatial and temporal dynamics of muscle activity \cite{mao2025syniemg}. Although these machine learning (ML) approaches have shown significant promise, they are purely data-driven and do not account for musculoskeletal variations in muscle anatomy and neural structures. This lack of physiological integration limits their transparency, reduces physiological accuracy, and hinders generalizability.

To address these challenges, one practical and efficient approach is to generate unmeasurable EMG signals from joint kinematic data by integrating the data-driven ML approach with the physics of a personalized musculoskeletal model.
A similar strategy exists in physics-informed neural networks (PINNs) \cite{karniadakis2021physics,meng2022physics,navaneeth2024physics}, where the features of physical systems are encoded in the ML models using governing equations of underlying physical systems with or without labeled data.
The success of combining data-driven learning with governing physical principles in musculoskeletal (MSK) modeling can already be observed for estimating muscle forces and joint kinematics from surface EMG signals. Various approaches, including CNN-based physics-informed models, physics-informed parameter identification neural networks, and GAN-based methods, have demonstrated enhanced accuracy in MSK modeling \cite{zhang2022physics,zhang2022boosting, taneja2022feature,shi2023physics,ma2024physics}. However, the use of PINN-based frameworks for inverse simulations of EMG signals of surface and deep muscles from joint kinematics has not been investigated in the literature.

To that end, in this paper, for the first time, we propose a physics-integrated deep learning algorithm that integrates the non-trivial concepts of physics-informed and data-driven deep learning to approximate the EMG envelope from the deeper muscles. By embedding physical laws from the MSK model, the proposed physics-integrated deep learning algorithm can efficiently learn the complex, generalizable features of high-dimensional MSK systems.
While physics-based modeling is used to predict the missing EMG envelope, data-driven modeling engraves the subject-specific information into the predictions.
In addition, they are well suited for personalized modeling as they adapt to individual anatomical differences with limited data, making them particularly effective for constructing deep muscle EMG signals. 
The contributions of the proposed framework to the existing literature can be encapsulated into the following three points:
\begin{enumerate}
    \item[\textbf{I.}] \textbf{Construction of deep muscle EMG:}  
    We propose a novel neural musculoskeletal model (NMM) capable of constructing non-measurable deep muscle EMG signal envelopes entirely from joint kinematic states (angle and velocity) and lifting load information in the upper limb.
    \item[\textbf{II.}] \textbf{Development of a personalized Musculoskeletal model:}  
    We designed a physics-integrated neural network architecture trained jointly using measurable joint kinematic states, surface EMG (sEMG) envelopes, and the physics of a musculoskeletal forward dynamics model. The model includes tunable parameters to ensure subject-specific personalization during training.
    \item[\textbf{III.}] \textbf{Generalization across multiple loading conditions:}
    A subject-specific NMM can be pre-trained and tested on multiple loading conditions. This eliminates the necessity for pre-training multiple ML models for different loading conditions, saving computational resources and the cost of pre-training multiple models from scratch. 
\end{enumerate}

The proposed NMM is validated in an experimentally controlled environment by predicting EMG envelopes across various movement and loading scenarios for five test subjects. The constructed EMG envelope are validated by predicting joint torques under diverse muscle loading conditions and comparing these predictions with inverse dynamics-based torque. For comparison, the constructed EMG signals are also compared with the state-of-the-art muscle synergy extrapolation (MSE) approach. The proposed NMM outperforms the existing MSE framework by a significant margin.

The rest of this article is organized as follows: Section \ref{sec:backgr} provides background and formulates the problem, including the forward dynamics of the musculoskeletal model. Section \ref{sec:metho} details the proposed methodology, covering the physics-informed neural network, customized loss function, various hyperparameters, and the training procedure. Section \ref{sec:valid} explains the system validation process, including the experimental setup, data collection, data preprocessing, and model evaluation criteria. Section \ref{sec:results} presents the results and the performance of the model, while Section \ref{sec:discussion} offers further discussion, and Section \ref{sec:conclusion} concludes the study.
\section{Background and problem formulation} \label{sec:backgr}
\subsection {EMG-based musculoskeletal forward dynamics model}
In this study, we use a generic EMG-driven musculoskeletal (MSK) model to simulate forward dynamics, integrating it with a neural network to construct deep muscle EMG signals. The generic MSK model incorporates Hill-type muscle actuators to represent individual muscle actuation, utilizing EMG signals and muscle geometry as inputs to generate force. Designed with scalable parameters and generalizability to various movements, this model can be calibrated to individuals through optimization. It consists of four primary modules: (A) the "muscle activation dynamics module," (B) the "muscle contraction dynamics module," (C) the "musculotendon kinematics module," and (D) the "joint torque computation module," as detailed below.
\subsubsection{Muscle activation dynamics module}
In this module, EMG signals are converted into muscle activation \cite{milner1973changes,zajac1989muscle,buchanan2004neuromusculoskeletal,sartori2012emg}. The module first transforms EMG signals into muscle excitation $u(t) \in [0, 1]$  and then into muscle activation $a(t) \in [0, 1]$. Given that EMG data are sampled discretely, mapping EMG data to muscle excitation is represented by the recursive relation in Eq. \eqref{muscle_excitation} (see \cite{buchanan2004neuromusculoskeletal} for more details)
\begin{equation}\label{muscle_excitation}
u_j[n] = \alpha e_j[n-d] - \beta_1 u_j[n-1] - \beta_2 u_j[n-2]
\end{equation}
where $e_j$ is the filtered EMG envelope, and $d \in \mathbb{R}$, $\alpha \in \mathbb{R}$, $\beta_1 \in \mathbb{R}$, and $\beta_2 \in \mathbb{R}$ are parameters related to muscle excitation, which are subjected to the following constraints.
\begin{equation}\label{constraint one}
\begin{aligned}
    & (i) \left|c_1\right| < 1, & (ii) \left|c_2\right| < 1, \\
    & (iii) \beta_1 = c_1 + c_2, & (iv) \beta_2 = c_1 \cdot c_2, \\
    & (v) \alpha - \beta_1 - \beta_2 = 1 &
\end{aligned}
\end{equation}
The muscle activation $a[n]$ is obtained as follows:
\begin{equation}\label{non-linear relationship between stimulation frequency and activation1}
    a_j[n] = \frac{e^{A_j u_j[n]} - 1}{e^{A_j} - 1}
\end{equation}
where parameter $A_j \in [-3, 0]$ is a non-linear shape factor; a value of $-3$ indicates highly exponential behavior and $0$ indicates a linear behaviour of the activation.
\subsubsection {Musculotendon kinematics module}
The musculotendon kinematics module computes musculoskeletal geometry, including muscle lengths (MLs) and moment arms (MAs) of each muscles across joint angles $q_i$. For upper limb muscles, the polynomial equations in Eq. \eqref{muscle length regression equation} provide the relationship between muscle length, moment arm, and joint angle \cite{pigeon1996moment} for $i=\{1,2\}$ joints. 
\begin{equation}\label{muscle length regression equation}
    \begin{aligned}
        \text{ML}_j[n] &= \kappa_j + \sum_{i=1}^2\left(y_n q_i^n+y_{n-1} q_i^{n-1}+\cdots+y_1 q_i\right) \\
        \text{MA}_j^{\text{dof}}[n] &= r_j + x_n q_i^n+x_{n-1} q_i^{n-1}+\cdots+x_1 q_i
    \end{aligned}
\end{equation}
where $\text{ML}_j$, and $\text{MA}_j$ denote the length and moment arm (at a specific degree of freedom (dof)) for the $j^{th}$ muscle in millimeters, respectively. The term $q_i$ represents the joint angle in degrees. The parameters $\kappa_j$ and $r_j$ represent the constant length of the muscle and moment arm, respectively.
Further, $y_i$ and $x_i$ denote the coefficients for various upper limb movements, such as elbow flexion and extension, as presented in \cite{pigeon1996moment}.

\subsubsection{Muscle contraction dynamics module}
The muscle contraction dynamics module uses muscle activations $a[n]$ and muscle lengths $\text{ML}_j[n]$ ($l^m[n]$ from here on) to determine the muscle force ($F^m[n]$) produced by each muscle at $n^{th}$ time step. 
The Hill-type muscle model utilized here \cite{zajac1989muscle,sartori2012emg,lloyd2003emg} consists of two main components: the contractile element (CE) and the parallel elastic element (PE). These components account for the contractile $F^{\text{CE}}[n]$ and passive forces $F^{\text{PE}}[n]$ produced by the muscle fibers, respectively. 
The contractile force in the muscle is computed as
\begin{equation}
    F^{\text{CE}}[n]=F_0^m a[n] \bar{f}_l^{\text{a}}\left(\bar{l}^m[n]\right) \bar{f}_v\left(\bar{v}^m[n]\right)
\end{equation}
where $\bar{f}_l^a(\bar{l}^m[n])=\exp({-(\bar{l}^m[n]-1)^2 / \gamma})$ is a Gaussian function representing the normalized active force-length relationship for all muscles in the musculoskeletal model. $F_0^m$, $\bar{l}^m[n]$, and $\gamma$ denote the maximum isometric force, normalized muscle fiber length, and a shape factor for the Gaussian active force-length relationship, respectively. The $\bar{f}_v\left(\bar{v}^m[n]\right)$ normalized force-velocity relationship is defined as
\begin{equation}\label{Eqn: force velocity}
    \bar{f}_v(\bar{v}^m[n])=
    \left\{\begin{aligned}
    &\frac{(\bar{v}^m+1)}{1+{A_f^{-1}}}; \quad \bar{v}^m<-1 \\
    &\frac{(\bar{v}^m+1)}{1-{\bar{v}^m}{A_f^{-1}}}; \quad-1 \leq \bar{v}^m<0 \\
    &\frac{\left(2+{2}{A_f^{-1}}\right) \bar{v}^m \bar{F}_{\text {len }}^M+\bar{F}_{\text {len }}^M-1}{\left(2+{2}{A_f^{-1}}\right) \bar{v}^m+\bar{F}_{\text {len }}^M-1}; \quad 0 \leq \bar{v}^m < \psi \\
    &\frac{\bar{F}_{\text {len }}^M}{20\left(\bar{F}_{\text {len }}^M-1\right)}  \left(\frac{\left(1+A_f^{-1}\right) \bar{F}_{\text {len }}^M \bar{v}^m}{10\left(\bar{F}_{\text {len }}^M-1\right)}+ 18.05 \bar{F}_{\text {len }}^M-18 \right), \; \bar{v}^m > \psi
    \end{aligned}\right.
\end{equation}
where $\psi$ is given as 
\begin{equation}
    \psi = \frac{10\left(\bar{F}_{\text {len }}^M-1\right)\left(0.95 \bar{F}_{\text {len }}^M-1\right)}{\left(1+A_f^{-1}\right) \bar{F}_{\text {len }}^M}
\end{equation}
and the symbols $\bar{v}^m$, $A_f$, and $\bar{F}_{\text {len }}^M$ are normalized muscle fiber length contraction rate, a shape factor for the force-velocity relationship, and maximum normalized muscle force achievable when the fiber is lengthening, respectively. 

The passive force in the muscle fibers is computed as 
\begin{equation}
    F^{\text{PE}}[n]=F_0^m \bar{f}_l^{\text{p}}\left(\bar{l}^m[n]\right)
\end{equation}
where $\bar{f}_l^{\text{p}}(\bar{l}^m[n])$ the normalized passive force-length relationship is defined as
\begin{equation}\label{Eqn: passive force len}
    \bar{f}_l^{\text{p}}\left(\bar{l}^m[n]\right) = 
    \left\{\begin{aligned}
        &1+\frac{k^{\text{PE}}}{\varepsilon_0^M}\left(\bar{l}^m-\left(1+\varepsilon_0^M\right)\right); \bar{l}^m>1+\varepsilon_0^M \\
        &\frac{\exp\left({k^{\text{PE}}\left(\bar{l}^m-1\right) \varepsilon_0^M}\right)}{\exp\left({k^{\text{PE}}}\right)}; \;\; \bar{l}^m \leq 1+\varepsilon_0^M
    \end{aligned}\right.
\end{equation}
where the symbols $k^{\text{PE}}$ and $\varepsilon_0^M$ represent an exponential shape factor for the passive force-length relationship and passive muscle strain due to maximum isometric force.

Finally, to compute the muscle force, the contraction dynamics model is represented by the following equation,
\begin{equation}\label{Muscle contraction dynamics}
    F_j^m[n]=\left[F_j^{\text{CE}}[n]+F_j^{\text{PE}}[n]\right] \cos (\phi_j[n])
\end{equation}
where $F_j^m[n]$ is the force exerted by the $j^{th}$ muscle at the $n^{th}$ time stamp. $\phi_j[n]$ denote pennation angle at the $n^{th}$ time stamp \cite{thelen2003adjustment}. The pennation angle is defined as
\begin{equation}\label{pennation angle}
    \phi_j[n] = \left\{\begin{aligned}
                        &0; & l^m=0\, \text{or} \, w \leq 0 \\
                        &\sin^{-1}\left(w\right); & 0<w<1 \\
                        &\pi / 2; & w \geq 1
                        \end{aligned}\right.
\end{equation}
where $w \in \mathbb{R}$ and $l^M \in \mathbb{R}$ denotes the muscle fibre width and length, respectively.  

\begin{equation}\label{width muscle}
    w_j[n]=\frac{L_j^o \sin (\phi_o)_j}{l_j^m}
\end{equation}
where ${L_j^o}$, $(\phi_o)_j$ and $l_j^m$ are optimal fiber length, optimal pennation angle, and muscle length of the $j^{th}$ muscle, respectively. It is to be noted that in the muscle contraction dynamics module, parameters $F_0^m$ and ${L_j^o}$ are subject-specific and need to be calibrated.

\subsubsection{Joint torque computation module}
This module computes joint torque through interaction between the "musculo-tendon kinematics module" and the "muscle contraction dynamics module." In particular, this is achieved by adding up the product of the muscle force $F_j^m[n]$ and the moment arm $\text{MA}_j^{\text{dof }}[n]$ for each ${\text{dof }}$ at $n^{th}$ time stamp, as follows.
\begin{equation}\label{Equation torque computation}
    \tau^{\text{dof }}[n] = \sum_{j=1}^{\text{k}}F_j^m[n] \cdot \text{MA}_j^{\text{dof }}[n]
\end{equation}

\subsection{Problem formulation}
In the above musculoskeletal forward dynamics model, the computation of the elbow joint torque $\tau^{\mathrm{dof}}$ in the upper limb requires the measurement of the EMG signals $\{e_j; j=1,\ldots,6\}$ (see Eq. \eqref{muscle_excitation}) at all six muscles. \textit{While the EMG envelopes at the four muscles, biceps long head (Biclong), biceps short head (Bicshort), triceps long head (Trilong), and triceps lateral head (Trilat), are possible to measure in a non-invasive manner using sEMG sensors, the EMG signal envelopes at the deeper muscles, brachialis (Brach) and triceps medial head (Trimed), are either constructed or measured using invasive methods or not included in the measurement}.
Since the musculoskeletal forward dynamics model is subject-specific, obtaining the deeper or missing muscle EMG signals creates a bottleneck in the global adoption of the above musculoskeletal model. Thus, it becomes necessary to have a numerical framework that can provide subject-specific EMG envelops of the non-measurable muscles. 

In this work, we proposed a neural musculoskeletal model (NMM) that can construct the non-measurable EMG signal envelopes of the \textit{brachialis} and \textit{triceps medial head} muscles entirely from the information of joint kinematic state and lifting loads. The underlying problem can be expressed as,
\begin{equation}
    \{e_b,e_t\} = \mathcal{N}\left( \bm{q}, \dot{\bm{q}}, m, e_{1:4}; \bm{\theta} \right),
\end{equation}
where $\mathcal{N}_{\theta}$ is a neural network parameterized by the tunable parameters $\bm{\theta}$, $e_b$ and $e_t$ are the EMG signals of \textit{brachialis} and \textit{triceps medial head} muscles, $\bm{q}$ and $\dot{\bm{q}}$ are the joint angle and angular velocity of the joint, $m$ is the mass of lifted weight, and $e_{1:4}$ are the EMG signals of measurable surface muscles. 
The network parameters are optimized by minimizing the loss between the inverse dynamics model and the predicted torque from the trained musculoskeletal forward dynamics model,
\begin{equation}
    \bm{\theta}^* = \underset{\theta}{\arg \min} \; \mathcal{L}\left(\tau(\bm{q},m), \tau^p(e_b,e_t,e_{1:4},\bm{q}) \right)
\end{equation}
where $\tau(\bm{q},\dot{\bm{q}},m)$ is the torque from the inverse dynamics model and $\tau^p(e_b,e_t,e_{1:4},m)$ is the predicted torque from the trained musculoskeletal forward dynamics model.

\begin{figure*}[t!]
    \centerline{\includegraphics[trim=0.08cm 0.2cm 0.2cm 0.4cm, clip=true,width=\textwidth]
    {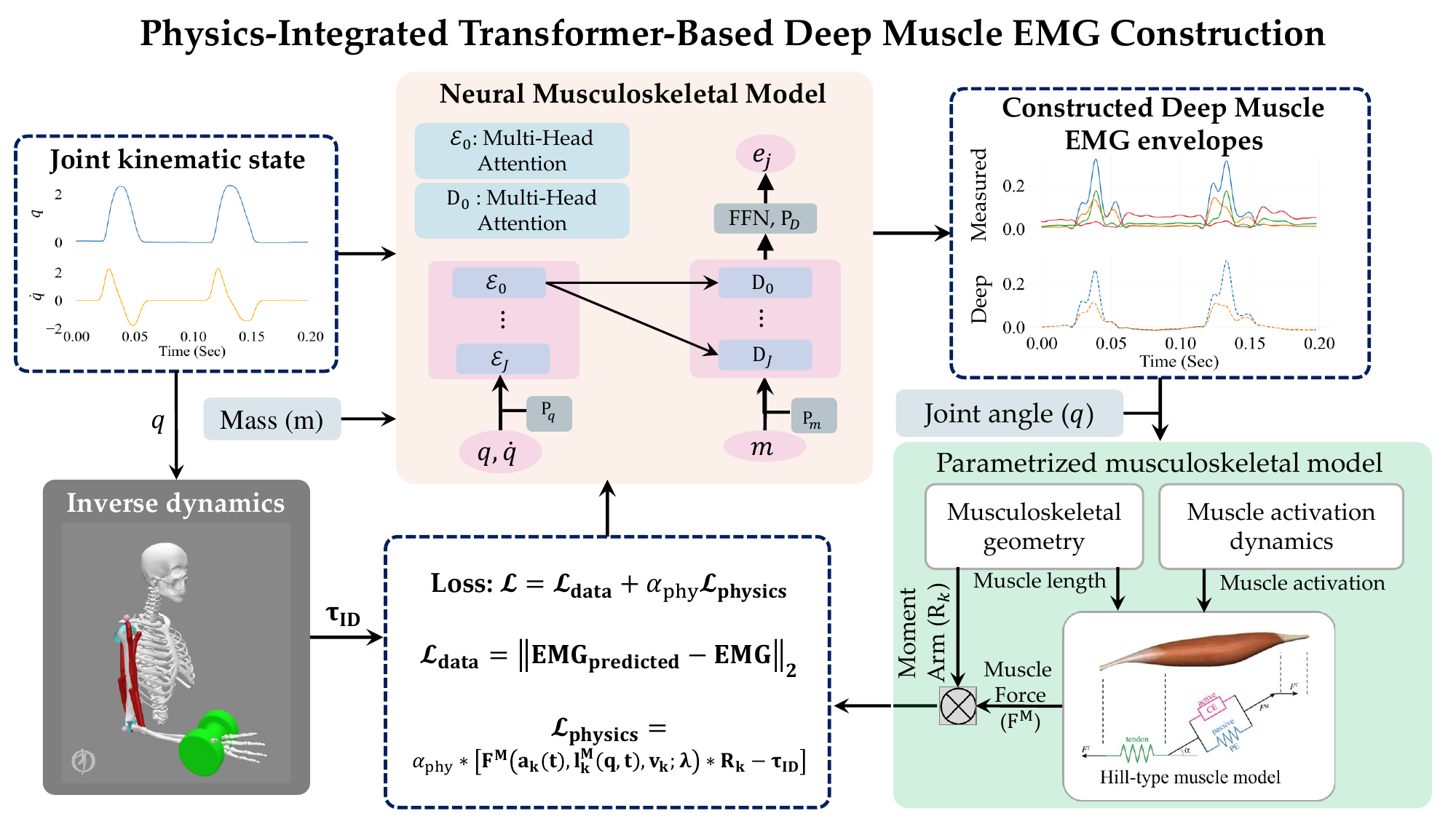}}
    \caption{Schematic architecture of the proposed Neural Musculoskeletal Model for constructing EMG signals from joint kinematics.}
    \label{Fig:PiGRN}
\end{figure*}
\section{Proposed methodology} \label{sec:metho}
We model the neural network $\mathcal{N}_{\theta}$ as an attention-transformer. The transformer architecture has an encoder-decoder structure, wherein the encoder-decoder block uses a series of attention encoders and decoders. The encoder takes the joint kinematic state as input, whereas the decoder takes the mass of weights as input. 
The transformer architecture predicts EMG envelopes for six muscles, four measurable and two non-measurable. A composite loss is calculated over the predicted EMG signals. As a part of the composite loss, the data loss is calculated on the four measured EMG envelopes, and a physics loss is calculated using the musculoskeletal forward dynamics of the six predicted EMG envelopes.
By coupling the $\mathcal{N}_{\theta}$ with physics from the musculoskeletal forward dynamical model, we construct a physics-integrated neural network architecture. 
The framework of the proposed neural musculoskeletal model (NMM) is illustrated through a schematic diagram in Fig. \ref{Fig:PiGRN}.

Once trained, the proposed neural musculoskeletal model constructs six EMG signal envelopes. However, the constructed EMG envelopes corresponding to measurable muscles are discarded, and only the two non-measurable EMG envelopes are taken as output. The details of the transformer architecture are provided in the following section.

\subsection{Physics-integrated neural network (PINN)}
We consider a dataset $\mathcal{D} = \{\mathcal{I}, \mathcal{O}\}$ comprising the input $\mathcal{I}$ and output $\mathcal{O}$, where the input-output pairs are defined as $\mathcal{I}= \{q_1(t), q_2(t), m\}$ and $\mathcal{O} = \{e_j(t); 1, \ldots, 6\}$. 
We define the encoder and decoder inputs as $\mathcal{I}_{q}= \{q_1(t), q_2(t)\}$ and $\mathcal{I}_{m}= m$ which are a subset of the input set $\mathcal{I}$.
In the encoder, the input $\mathcal{I}_{q}$ is first projected to a high-dimensional space as $\bm{\vartheta}_0 = \mathrm{P}_{q}(\mathcal{I}_{q})$. The transformation $\mathrm{P}_{q}: \mathbb{R}^{d_{\mathcal{I}_{q}}} \mapsto \mathbb{R}^{d_l}$ is modeled using a linear feed-forward network, where $d_l$ is the dimension of the lifted space. 
We update the lifted input $\vartheta_0$ through $J$ number of attention encoders, expressed as,
\begin{equation}
    \vartheta_{J}(t) = \left(\mathcal{E}_{J} \circ \mathcal{E}_{J-1} \circ \ldots \circ \mathcal{E}_0 \right) \left(\bm{\vartheta}_0 \right)(t),
\end{equation}
where $\mathcal{E}_j \in \mathbb{R}^{d_{l_j}} \mapsto \mathbb{R}^{d_{l_{j+1}}}$ denotes the $j^{th}$ encoder. This encoder is modeled as multi-head attention blocks, which obtain the key, query, and values from the lifted encoder input.

Similarly, the decoder input $\mathcal{I}_{m}$ is first projected into a high-dimensional space through a linear transformation $\mathrm{P}_{m}: \mathbb{R}^{d_{\mathcal{I}_{m}}} \mapsto \mathbb{R}^{d_l}$, which is again modeled using a linear feed-forward network with $d_h$ denoting the dimension of the lifted space. Note that we consider the dimension of the high-dimensional space to be the same in both the encoder and decoder. 
The desired EMG signal is obtained by iteratively updating the lifted decoder input $\bm{\eta}_0 = \mathrm{P}_{m}(\mathcal{I}_{m})$ using $J$ a number of cross-attention decoders. We define the decoders as $\mathrm{D}_j \in \mathbb{R}^{d_{l_j}} \mapsto \mathbb{R}^{d_{l_{j+1}}}$ as multi-head attention blocks that obtain the query from $\bm{\eta}_0$ and the key and values from the encoder output $\bm{\vartheta}_{J}(t)$. 
The iterations of the decoder are expressed as,
\begin{equation}
    \bm{\eta}_{j+1}(t) = \mathrm{D}_j \left(\bm{\eta}_j, \bm{\vartheta}_{J} \right)(t), \;\; j\in[1,J]
\end{equation}
where note that all the decoder blocks use the final encoder output $\bm{\vartheta}_{J}(t)$. The iteratively updated output $\bm{\eta}_J(t) \in \mathbb{R}^{d_{l_J}}$ is projected to six EMG signals using a dense feed-forward network. We denote the projection as $\mathrm{P}_{D}: \mathbb{R}^{d_{l_J}} \mapsto \mathbb{R}^6$, which in our case is modeled as a three-layered feed-forward network with GeLU activation function. 
The feed-forward network has a sigmoid activation function at the output, defined as $\sigma(x) = 1/(1+ \exp(-x))$, which restricts the range of EMG envelopes between 0 and 1. The embedding dimension in both the encoder and decoder is kept as 6.

\subsection{Encoder and decoder blocks}
In the above network architecture, we utilize multi-head attention blocks \cite{vaswani2017attention} as encoders and decoders. Each encoder $\mathcal{E}_j$ is composed of two sub-layers, the first involving a multi-head self-attention mechanism (MultiHead), while the second involving a dense feed-forward network (FFN). In particular, the operation inside an encoder is expressed as follows,
\begin{equation}
    \begin{aligned}
        \bm{\vartheta}_{j+1}(t) &= (\mathrm{SubLayer}_1 \circ \mathrm{SubLayer}_2)(\bm{\vartheta}_{j})(t) \\
        \mathrm{SubLayer}_1 &= \mathrm{LayerNorm}(\bm{\vartheta}_{j} + \mathrm{MultiHead}(\bm{Q}=\bm{\vartheta}_{j},\bm{K}=\bm{\vartheta}_{j},\bm{V}=\bm{\vartheta}_{j})) \\
        \mathrm{SubLayer}_2 &= \mathrm{LayerNorm}(\mathrm{SubLayer}_1 + \mathrm{FFN}(\mathrm{SubLayer}_1))
    \end{aligned}
\end{equation}
where $\bm{Q}$, $\bm{K}$, and $\bm{V}$ denote the query, key, and value of the attention mechanism. Given the query, key, and value weight matrices, $\mathbf{W}_i^Q \in \mathbb{R}^{d_h \times d_k}$, $\mathbf{W}_i^K \in \mathbb{R}^{d_h \times d_k}$, and $\mathbf{W}_i^V \in \mathbb{R}^{d_h \times d_v}$, the multi-head attention is performed as,
\begin{equation}
    \mathrm{MultiHead}(\bm{Q},\bm{K},\bm{V}) = \Lambda\left( \mathrm{Concat}(\mathcal{A}_1, \ldots, \mathcal{A}_h) \right)
\end{equation}
where $\mathcal{A}_i$ is the scaled dot-product attention and $\Lambda: \mathbb{R}^{d_h d_v} \mapsto \mathbb{R}^{d_l}$ is a projection that maps the concatenated dimension back to the embedding dimension. Further note that $d_h$ is the number of attention heads, $d_k$ is the dimension of key, and $d_v$ is the dimension of value. The scaled dot-product attention is defined as
\begin{equation}
    \mathcal{A}_i(\bm{Q},\bm{K},\bm{V}; \mathbf{W}_i^Q, \mathbf{W}_i^K, \mathbf{W}_i^V) = \mathrm{Softmax}\left( \frac{(\bm{Q} \mathbf{W}_i^Q)(\bm{K} \mathbf{W}_i^K)^{\top}}{\sqrt{d_k}} \right) (\bm{V} \mathbf{W}_i^V)
\end{equation}

The decoder structure is similar to the encoder, with the addition of an extra sub-layer positioned between the self-attention and FNN sub-layers. This intermediate layer is a multi-head attention block that performs cross-attention, linking the output of the encoder stack with the decoder input. This is different from the self-attention mechanism in the encoder in the way that it obtains the query from the decoder input, whereas all the query, key, and values are obtained from the encoder input in the self-attention. 
The operation inside a decoder is expressed as follows,
\begin{equation}
    \begin{aligned}
        \bm{\eta}_{j+1}(t) &= (\mathrm{SubLayer}_1 \circ \mathrm{SubLayer}_2 \circ \mathrm{SubLayer}_3)(\bm{\eta}_{j})(t) \\
        \mathrm{SubLayer}_1 &= \mathrm{LayerNorm}(\eta_{j} + \mathrm{MultiHead}(\bm{Q}=\bm{\eta}_{j},\bm{K}=\bm{\eta}_{j},\bm{V}=\bm{\eta}_{j})) \\
        \mathrm{SubLayer}_2 &= \mathrm{LayerNorm}(\mathrm{SubLayer}_1 + \mathrm{MultiHead}(Q=\mathrm{SubLayer}_1,K=\bm{\vartheta}_{J},V=\bm{\vartheta}_{J})) \\
        \mathrm{SubLayer}_3 &= \mathrm{LayerNorm}(\mathrm{SubLayer}_2 + \mathrm{FFN}(\mathrm{SubLayer}_2))
    \end{aligned}
\end{equation}
In the above operation, note that the cross-attention layer in the decoder takes the final output of the encoder stack. 
The architecture of the FFN across all the encoders and decoders is the same with different parameters. In this work, we utilize a seven-layered FFN with the perceptions $[6,64,128,256,128,64,6]$, where the number of perceptions in the first layer indicates the embedding dimension.

\subsection{Customized loss function and training}
The network parameters and the physical parameters of the NMM are optimized using the following loss functions,
\begin{equation}
    \mathcal{L} = \mathcal{L}_{\mathrm{data}} + \lambda \mathcal{L}_{\mathrm{phy.}},
\end{equation}
where $\mathcal{L}_{\mathrm{data}}$ and $\mathcal{L}_{\mathrm{phy.}}$ represent the data and physics loss, respectively. The constant $\lambda \in \mathbb{R}$ is the participation factor of the physics loss. While $\mathcal{L}_{\mathrm{data}}$ helps the NMM to predict the unmeasurable EMG in a way that the contribution from the measured EMG signals is preserved, the $\mathcal{L}_{\mathrm{phy.}}$ guides the unmeasurable EMG predictions to satisfy the torque requirements. 
These losses are defined as, 
\begin{equation}\label{eq:combined_loss}
    \begin{aligned}
        \mathcal{L}_{\mathrm{data}} &= \frac{\sqrt{\sum_{i=1}^{n} \sum_{j=1}^{4} \left(e^{*}_{ij} - e_{ij} \right)^2}}{\sqrt{\sum_{i=1}^{n} \sum_{j=1}^{4} (e^{*}_{ij})^2}} \\
        \mathcal{L}_{\mathrm{phy.}} &= \frac{\sqrt{\sum_{i=1}^{n} \left(\tau^{*}_{i} - \tau_{i} \right)^2}}{\sqrt{\sum_{i=1}^{n} (\tau^{*}_{i})^2}} 
    \end{aligned}
\end{equation}
where $e_{ij}$ and $e^{*}_{ij}$ are the true and predicted EMGs, and $\tau_{i}$ and $\tau^{*}_{i}$ are the true and predicted EMGs. While the true EMGs are recorded experimentally from the surface muscles, the true torque is inferred from the inverse dynamics model, which in our case is the `OpenSim' software \cite{delp2007opensim}. 
By minimizing the composite loss $\mathcal{L}$, the network parameters are optimized.

\subsection{Personalized Musculoskeletal parameters}
We present a physics-integrated neural musculoskeletal model that allows subject-specific personalization during training and testing.
To allow the proposed NMM to adapt to subject-specific features, we introduce 32 tunable parameters in the musculoskeletal forward dynamics model.  
In addition to the neural network parameters $\bm{\theta}$, we optimize the 32 parameters (across six units of muscles) for each subject to allow the NMM to construct the EMG data for the deep muscles accurately. 
The 32 parameters include 6 nonlinear shape parameters $\{A_j;j=1,\ldots,6\}$, 2 muscle activation constants $\{c_1,c_2\}$, 6 maximal isometric force parameters $\{(F_o^{\operatorname{m}})_j;j=1,\ldots,6\}$, and 6 optimal fiber length parameters $\{L^o_j;j=1,\ldots,6\}$, 6 muscle length constants, $\{\kappa_j;j=1,\ldots,6\}$ and 6 moment arm constants $\{r_j;j=1,\ldots,6\}$.
\subsection {Hyperparameters of the model}
The linear layers $\mathrm{P}_q$ and $\mathrm{P}_m$ used for embedding the inputs in the encoder and decoder have 6 perceptions. The output projection network $\mathrm{P}_D$ at the end of the decoder is modeled as a three-layered dense network with a GeLU ($\sigma(x) = 0.5x \operatorname{erf}(x/\sqrt{2})$) nonlinear activation function between the layers and a sigmoid ($\sigma(x) = 1/1+\operatorname{exp}(-x)$) nonlinear activation at the output. The dense layers have 64, 32, and 6 perceptions, respectively. 
There are four encoder $\mathcal{E}_j$ and decoder $\mathrm{D}_j$ blocks in the transformer architecture. Each encoder and decoder transformer block share the configuration in Table \ref{tab:tarnsformer}.
The AdamW optimizer with the configuration provided in Table \ref{tab:optimizer} is used to optimize the neural network parameters.
The above configuration takes about 4 hours to complete 1000 epochs on a Xeon E-2236 CPU with 32GB memory.
While studying the performance of the proposed framework with the muscle synergy extrapolation (MSE) approach, the following parameters are considered.
\section{System validation procedures}\label{sec:valid}
\subsection{Muscle synergy extrapolation approach}
To provide a baseline comparison, we considered the muscle synergy extrapolation (MSE) approach. For all measured EMG data, represented as the matrix \( \mathbf{M} \in \mathbb{R}^{\text{len\_data} \times N_m} \), where \( \text{len\_data} \) is the number of time samples and \( N_m \) is the number of muscles, deep muscle EMGs were estimated using the MSE algorithm constrained with musculoskeletal (MSK)-based joint torque error minimization \cite{ao2020evaluation, tahmid2024upper}. The algorithm used non-negative matrix factorization (NMF) \cite{rabbi2020non} as a dimensionality reduction technique to decompose \( \mathbf{M} \) into a reduced number of time-varying synergy activations \( \mathbf{S} \in \mathbb{R}^{\text{len\_data} \times N_s} \) and time-invariant synergy weights \( \mathbf{W} \in \mathbb{R}^{N_s \times N_m} \), where \( N_s \) is the number of synergies. These time-varying activations were then used as basis functions to construct missing muscle EMG data. Mathematically, this is represented as:
\[
\mathbf{M} = \mathbf{S} \cdot \mathbf{W} + \mathbf{\epsilon},
\]
where \( \mathbf{\epsilon} \in \mathbb{R}^{\text{len\_data} \times N_m} \) represents the construction error or residual. Deep muscle EMGs were analyzed separately for each load condition (0, 2, and 4 kg) using the MSE algorithm. The extracted synergy activations successfully predicted the measured muscle EMG with a variance accounted for (VAF) exceeding 95\% when using three synergies \( N_s = 3\). The pseudo-code for the MSE algorithm is presented in Algorithm \ref{algo:MSE}. 

We implemented the MSE algorithm in Python with PyTorch. The model incorporates two learnable parameters: 
\(\mathbf{S} \in \mathbb{R}^{\text{len\_data} \times N_s}\), and 
\(\mathbf{W} \in \mathbb{R}^{N_s \times N_m}\). These parameters are optimized using the Adam optimizer and a mean squared error loss function. The key parameter values utilized in this implementation are detailed in Table \ref{tab:MASEparameters}.

\subsection{Experimental setup, data collection, and data processing}

\subsubsection{Experimental setup and data recording paradigm}
\begin{figure}[ht!]
    \centering
    \begin{subfigure}{0.48\textwidth}
        \includegraphics[trim=0.01cm 0.01cm 0.01cm 0.01cm, clip=true, width=\textwidth]{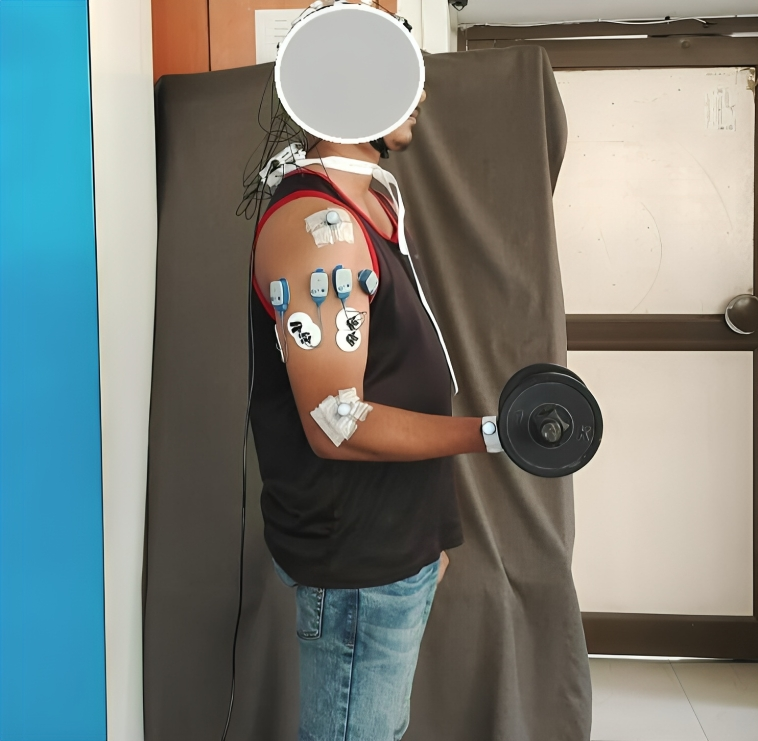}
        \caption{\textbf{Experimental setup for recording EMG and joint angle data}: Four EMG electrodes and three infrared reflective markers were strategically placed on the hand to capture muscle activity and joint movement. Participants performed elbow flexion and extension (eFE) movements while holding dumbbells of mass 0, 2, and 4 kg.}
        \label{fig:data_recording_setup}
    \end{subfigure}
    \hfill
    \begin{subfigure}{0.48\textwidth}
        \includegraphics[trim=0.01cm 0.01cm 0.01cm 0.01cm, clip=true, width=\textwidth]{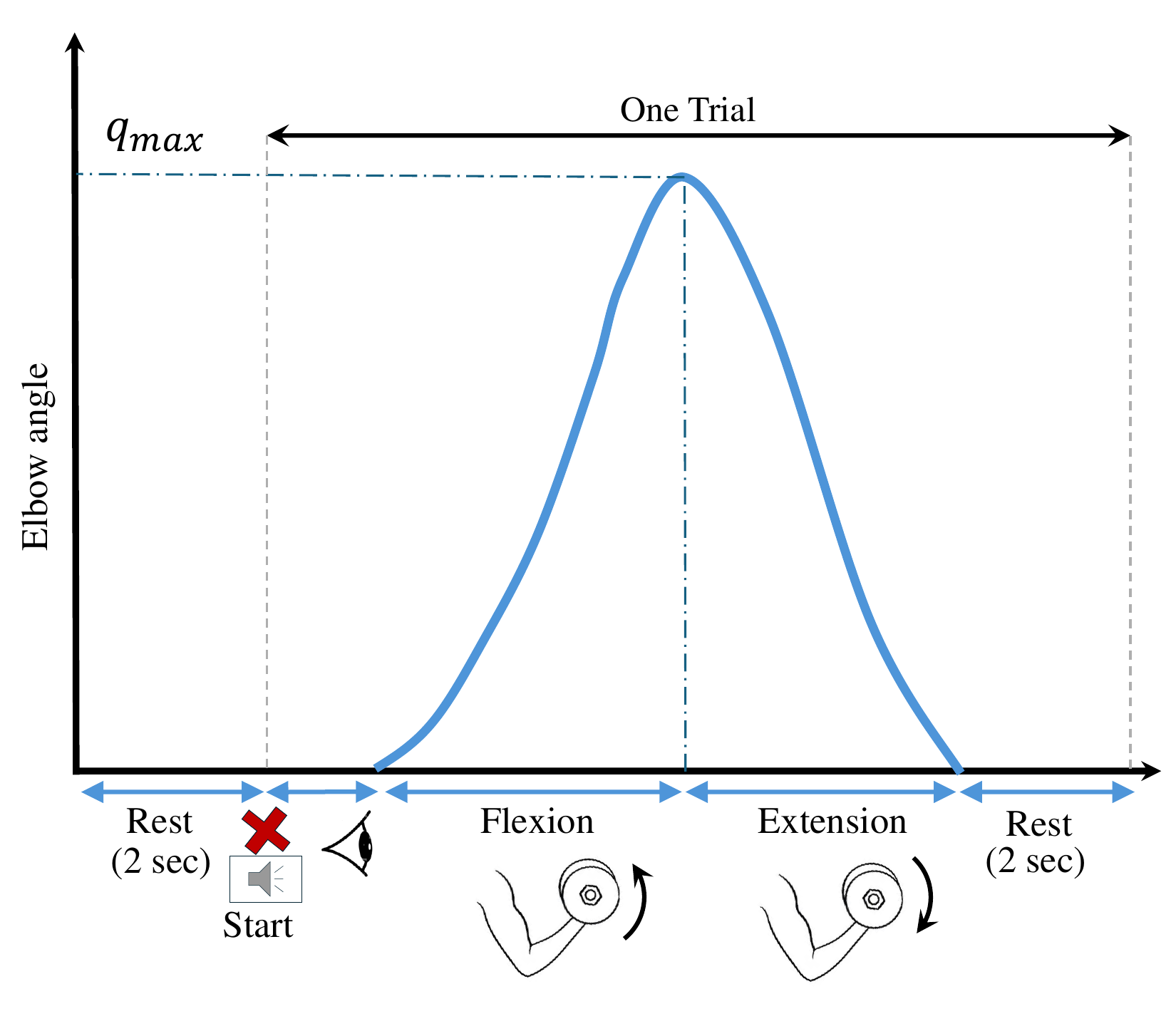}
        \caption{\textbf{Experimental Flow of the eFE Task}: The process consists of three phases – (1) Start Phase, where the participant stands with a balanced posture awaiting the cue and beep, (2) Execution Phase, during which the elbow flexion-extension (eFE) movement is performed, and (3) Resting Phase, where the participant holds the position for 2 seconds before the next trial.}
        \label{fig:additional_view}
    \end{subfigure}
    \caption{Illustration of the experimental setup used for data collection.}
    \label{fig:experimental_setup}
\end{figure}

The experiment involved five healthy right-handed male participants, aged $27 \pm 4$ years, with an average body mass of $70 \pm 6$\,kg and a height of $170 \pm 5.5$\,cm. Surface electromyography (sEMG) and joint angle data were collected simultaneously during the study. The experimental setup and protocols were approved by the Institutional Review Board of the All India Institute of Medical Sciences, New Delhi. 
Four electrodes were placed to capture EMG signals: one on each of the biceps long head, biceps short head, triceps long head, and lateral head, as illustrated in Fig. \ref{fig:data_recording_setup}. The EMG electrode placement and skin preparation followed established guidelines \cite{konrad2005abc}. 
EMG signals were recorded using Noraxon’s wireless Ultium EMG sensor system at a sampling rate of 4000 Hz. The elbow joint angle was measured using a marker-based camera system (Noraxon NiNOX 125 Camera System) positioned two meters away from the subject in the sagittal plane. The camera system was integrated with the Noraxon myoResearch platform (MR 3.16) to capture elbow flexion-extension (eFE) movements. During post-processing, the elbow joint angle was determined using the myoResearch software. Reflective markers were tracked with a three-point angle measurement tool to compute the 2-D joint angle from the video recordings. The joint angle data were sampled at 125 Hz, and synchronization of the sEMG and joint angle data was achieved using the Noraxon myoSync device.

During the experiment, both the sEMG and motion data were gathered as the participant performed the eFE task. At the start of the experiment, the participant stood while holding dumbbells weighing 0, 2, or 4 kg in their right hand. A monitor positioned two meters in front of them provided the experimental instructions. The experimental flow of the eFE task is presented in Fig. \ref{fig:experimental_setup}.
The participant maintained a balanced posture while holding the dumbbell. The experiment was designed using PsychoPy \cite{peirce2007psychopy} to instruct participants on initiating the eFE movement. Each trial started with a cross displayed at the screen and an accompanying beep to signal the beginning of the trial. After a short delay, a visual cue appeared, prompting the participant to begin the eFE movement. Movement was performed during the execution phase, followed by a two-second resting phase at the end of each trial. Before data collection, verbal instructions were provided to the participants. 
A total of six sets were recorded, with two sets recorded for each load condition (0 kg, 2 kg, and 4 kg). Each set comprised 10 eFE task trials, and rest periods were provided between runs to prevent muscle fatigue.
\subsubsection{Data Pre-processing}
To enhance the quality of the joint angle data, a Gaussian filter is applied with a standard deviation of 10
and a window size of 6. To obtain the angular velocity, the first-order derivative of the joint angle with respect to time was taken. For the sEMG data, a series of processing steps were implemented to obtain the EMG envelopes. First, the raw sEMG signals were subjected to band-pass filtering within the frequency range of 10-450 Hz. It eliminates the recorded low- and high-frequency noise. Then, a full-wave rectification process was performed to inverse the negative amplitude signal. Then, after 7 Hz, low-pass filtering \cite{sartori2012emg} was performed to smooth the signal to get the EMG envelope. Finally, the obtained smooth EMG envelopes were downsampled to 125 Hz to match the sampling frequency of the joint kinematics.
The outcome of these steps is linear envelopes that represent the amplitude of the sEMG signals. To normalize these envelopes, they are divided by the maximum voluntary contraction (MVC) EMG values obtained from the set of MVC recorded trials. This normalization process helps standardize EMG data across different trials and subjects.

\subsection{Deeper muscle EMG construction and its evaluation criteria}
The proposed framework aims to construct the EMG signal envelope of two deep muscles, namely \textit{Brachialis} and \textit{Triceps medial head}. Once the PINN model is trained, joint kinematic data from the test data set are provided as input to the model to predict the EMGs of these two deep muscles, along with four measured EMGs. Together, these six EMGs, combined with optimized musculoskeletal (MSK) parameters, serve as input to a subject-specific forward dynamics MSK model to estimate joint torque.

To thoroughly assess the performance of the NMM framework and the accuracy and reliability of the construction of deep muscle EMG envelopes, a comprehensive set of six evaluation criteria was used. Time-series line plots were employed to analyze temporal consistency and muscle activation patterns across movement conditions. Additionally, heat maps of the constructed EMG envelopes were generated to visualize the spatial distribution of muscle activation under various conditions. For a quantitative evaluation of accuracy, the following metrics were applied:  
\begin{itemize}
    \item Power spectral density (PSD) analysis of constructed EMG was performed to verify the preservation of frequency domain characteristics, particularly in the low-frequency range. To estimate the PSD, Welch’s method \cite{welch1967use} was applied by averaging over multiple overlapping segments of the signal. The PSD is defined as:
    \begin{equation}
    P_{xx}(f) = \frac{1}{K} \sum_{i=1}^{K} \frac{1}{N} \left| X_i(f) \right|^2
    \end{equation}
    \begin{equation}
        X_i(f) = \sum_{n=0}^{N-1} x_i[n] e^{-j 2 \pi f_n / N}
    \end{equation}
    where \( K \) denotes the number of overlapping segments, \( N \) represents the number of samples per segment, \( X_i(f) \) is the Discrete Fourier Transform (DFT) of the \( i^{\text{th}} \) segment, and \( x_i[n] \) refers to the segmented signal in the \( i^{\text{th}} \) window.
    \item The L2 norm was calculated to provide a quantitative measure of the overall signal strength of the constructed EMG envelope. It is defined as:  
    \begin{equation}\label{L2Norm}
        \text{L2 norm} = \sqrt{\sum_{i=1}^{n} \hat{y}_i^2}
    \end{equation}
    where \( \hat{y} \) represents the predicted EMG values. This scalar metric enables comparisons of muscle activation intensity between subjects, muscles, and loading conditions.
    \item To further evaluate the robustness of the proposed model, the relative absolute error (RAE) and the Pearson correlation coefficient (PCC) were calculated for estimated joint torque. The RAE measures the relative difference between predicted and ground truth torque, defined as:  
    \begin{equation}\label{RAE}
        \text{RAE} = \frac{\sum_{i=1}^n |y_i - \hat{y}_i|}{\sum_{i=1}^n |y_i - \bar{y}|}
    \end{equation}  
    where \( y_i \) denotes the ground truth joint torque from inverse dynamics for the \( i \)-th observation, \( \hat{y}_i \) is the corresponding joint torque predicted from constructed EMG envelopes as input to the MSK model, \( \bar{y} \) is the mean of the true joint torque values, and \( n \) is the number of observations.
    \item The PCC, which assesses the linear relationship between predicted and ground truth torque values, is given by:  
    \begin{equation}\label{CorrelationCoefficient}
        \text{PCC} = \frac{\sum_{i=1}^{n} (x_i - \bar{x})(y_i - \bar{y})}{\sqrt{\sum_{i=1}^{n} (x_i - \bar{x})^2 \sum_{i=1}^{n} (y_i - \bar{y})^2}}
    \end{equation}  
    where \( x_i \) and \( y_i \) are the predicted and ground truth torque values for the \( i \)-th observation, and \( \bar{x} \) and \( \bar{y} \) are their respective means.
\end{itemize}
Finally, the physiological accuracy of the constructed EMGs was demonstrated by analyzing the evolution of MSK parameters during EMG prediction.
\section{Results}\label{sec:results}
In this section, we assess the performance of the proposed neural musculoskeletal model (NMM) based on the evaluation criteria outlined earlier. The analysis begins with an examination of the training process of the NMM framework. Following this, the performance of the model is evaluated by analyzing the temporal patterns and spatial distribution of deep muscle activation. Frequency domain analyses are performed by power spectral density analyses. L2 norms are calculated to evaluate the magnitudes of the constructed EMGs. Finally, the robustness of the NMM and the physiological accuracy of the constructed EMG signals are investigated by evaluating the predicted joint torques and the physiological parameters of the MSK model. Additionally, the results of the NMM are compared with those of the state-of-the-art non-negative matrix factorization (NMF)-based MSE approach.

\subsection {Evaluation of the PINN training process}
\begin{figure}[!ht]
    \centering
    \includegraphics[width=0.6\textwidth]{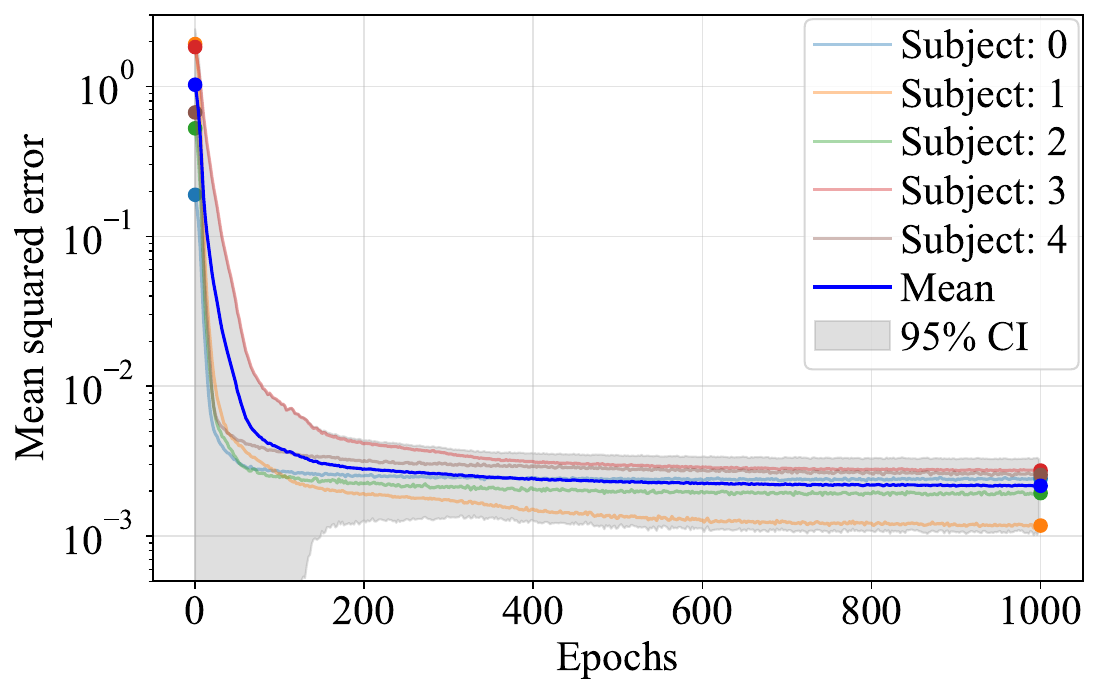}
    \caption{Training loss history of the proposed neural musculoskeletal model.}
    \label{fig:epoch_loss}
\end{figure}

The training loss history of the proposed NMM is illustrated in Fig. \ref{fig:epoch_loss}. During the training process, it is observed that both the training and testing losses decrease very rapidly within the initial epochs. However, the physical parameters of the NMM converge to a stationary value at a lower rate Fig. \ref{fig:parameters}. The effect of a sudden decrease in training loss in the initial few epochs may be attributed to the rapid decrease in data loss, whereas the physics loss regularizes the data loss at a slower rate. In such a scenario, the training is performed until both the composite loss (i.e., data loss + physics loss) and the physical parameter values converge to a stationary value, which in this case is found to be 1000 epochs. 

The evolution of the nonlinear shape parameter $A$, muscle activation constant $C$, maximal isometric force $F_o^{\operatorname{m}}$, and optimal fiber length, $L^o$ along with the muscle length $\kappa$ and moment arm $r$ constants for all six muscles, for a representative test subject, are illustrated in Fig. \ref{fig:parameters}. It is evident that as the training loss starts converging to a stationary value, the physical parameters also achieve a stationary value at the completion of the 1000th epoch.
Given the real-time observations of the joint angle $q$, its derivative $\dot{q}$, and the EMG observations from four muscles, the NMM provides the real-time predictions of the EMG signals for the third (\textit{Brach}) and sixth (\textit{Trimed}) muscles. It is also worth noting that the NMM provides the EMG predictions for different load levels.
\begin{figure*}[!ht]
    \centering
    \includegraphics[width= \textwidth]{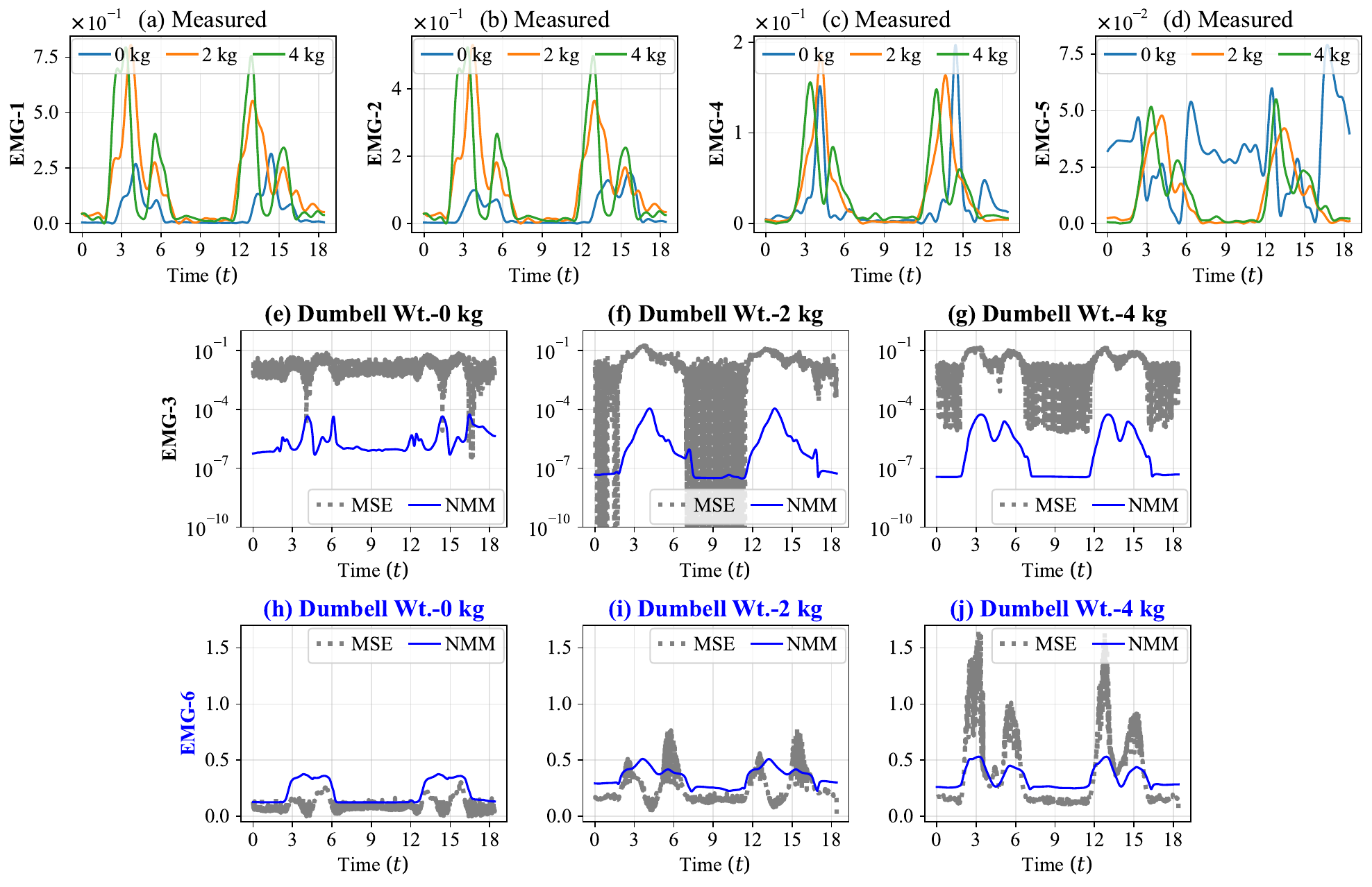}
    \caption{Description of the measured and constructed EMG envelopes at different loads. The EMG envelopes at the third and sixth muscles are constructed from the proposed neural musculoskeletal model, whereas EMG envelopes at other muscles are directly observed using sensors. The EMG signals at the third and sixth muscles at different loads are constructed using a single NMM.}
    \label{fig:emg_pinn}
\end{figure*}
\subsection{Temporal pattern in predicted EMG}
Fig. \ref{fig:emg_pinn} demonstrates the time-series line plots of the constructed EMG signal envelopes for the representative subject for two consecutive eFE movements. Here, plots (e, f, and g) and (h, i, and j) represent \textit{Brach} and \textit{Trimed} muscle EMG constructed using NMM, respectively, for lifted masses of 0, 2, and 4 kg, and their comparison with the MSE approach. Whereas the plots (a to d) are representative measured EMG envelopes for comparison purposes. The temporal consistency is clearly visible from these plots. The onset and offset of muscle activity, duration, and shape of activation bursts align with measured EMG envelope patterns.
\subsection{Spatial distribution pattern in predicted EMG}
\begin{figure*}[!ht]
    \centering
    \includegraphics[width=\textwidth]{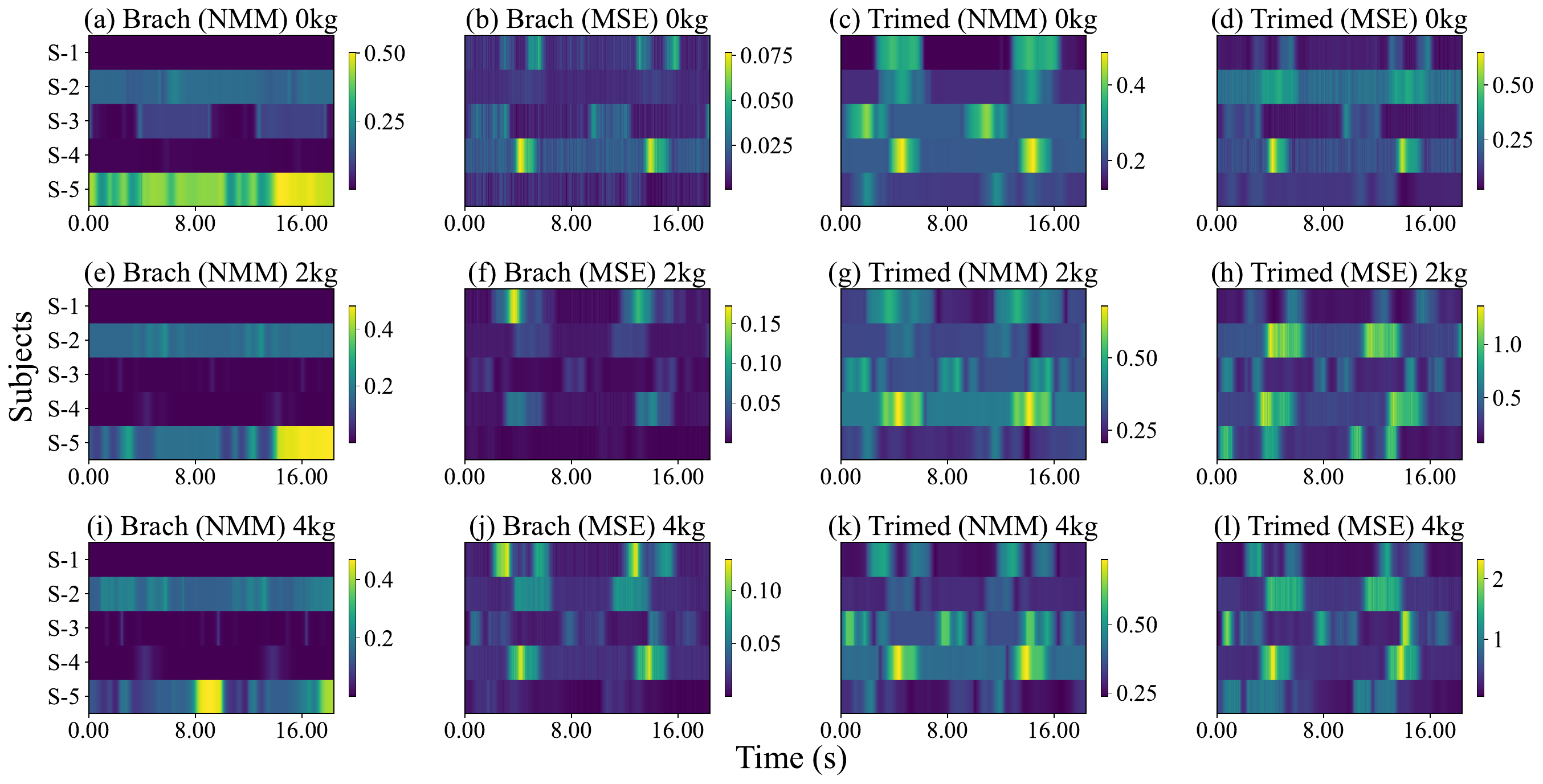}
    \caption{Heatmaps displaying the comparison of EMG profiles for the NMM and MSE methods across subjects (S-1 to S-5) and different load conditions (0 kg, 2 kg, and 4 kg). The plots represent two muscles: EMG2 (Brach) and EMG5 (Tri med) at each load. Each subplot provides insights into the temporal variations of muscle activation across subjects and methods under varying loads.}
    \label{fig:Heatmap}
\end{figure*}

Fig. \ref{fig:Heatmap} presents heat maps illustrating the spatial distribution of intensity of muscle activation with time across different loads (0 kg, 2 kg, and 4 kg) for \textit{Brach} and \textit{Trimed} using the NMM and MSE methods. Each row corresponds to a specific load, while the columns represent the results for the two muscles organized in a 3x4 grid. The y-axis denotes the subjects (S-1 to S-5), and the x-axis represents time in seconds. 
The intensity of the colors in the heatmaps reflects the magnitude of EMG activity, with dark or lighter colors indicating the moments of peak activation or muscle rest. The width of the lighter regions in the heat map provides an indication of the duration for which the muscle was active at higher levels of activation. For the \textit{Brach} muscle, EMG activity obtained from NMM and MSE both has dark colors throughout for all subjects except for the fifth subject, indicating very low activation values during eFE motion. 
However, the EMG data for the \textit{Trimed} muscle from both methods have two light-colored patches in between the dark, indicating muscle activation during two consecutive eFE motions for all subjects. Subject S-3 has performed three eFE motions within the same time duration, generating three light-colored patches. 
It should be noted that the activation level increases as the load increases, and the range of muscle activation obtained from NMM is between 0 and 1, whereas it goes beyond 1 from MSE. These results highlight the ability of the NMM to capture the spatial characteristics of muscle activation duration across different loads.
\subsection{Frequency domain analysis}
\begin{figure*}[!ht]
    \centering
    \includegraphics[width=\textwidth]{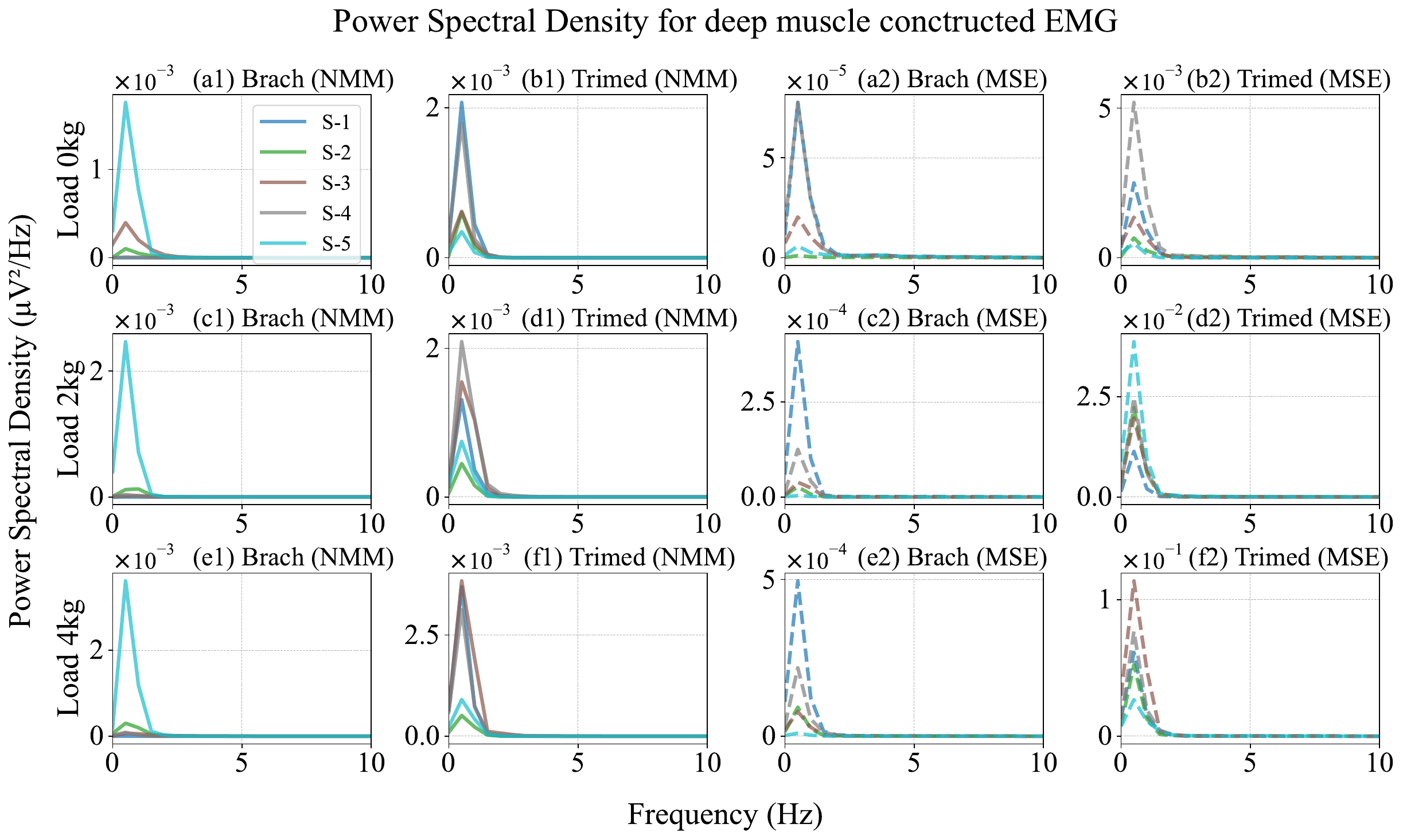}
    \caption{Power Spectral Density (PSD) of the two constructed EMG envelopes obtained from the NMM. The plot shows the constructed signal captures the low-frequency characteristics of the measured EMG, preserving essential neuromuscular dynamics.}
    \label{fig:PSD_recon}
\end{figure*}

The Power Spectral Density (PSD) is calculated to examine the frequency characteristics of the two constructed deep muscle EMG signal envelopes shown in Fig. \ref{fig:PSD_recon} for all five subjects.  
PSD values of two constructed EMG envelopes using NMM exhibit a single peak in a low frequency range $(<5\,\text{Hz})$ as presented in Fig. \ref{fig:PSD_recon} (a1 to f1). A first peak at low frequency indicates harmonic activation of the deep muscles with the measured muscle pairs \cite{boonstra2015muscle}. These results indicate that the deep muscle EMG envelope have the frequency domain characteristics. 
However, a notable difference in the magnitude of the PSD values is observed when comparing them with the EMG envelopes obtained using the MSE approach (Fig. \ref{fig:PSD_recon} a2 to f2), particularly at higher loads. For instance, the PSD value of \textit{Trimed} at a 4 kg load using NMM is on the order of \(10^{-3}\) \(\mu V^2/\text{Hz}\), compared to the corresponding order of \(10^{-1}\) \(\mu V^2/\text{Hz}\) in the MSE approach.

\subsection{Magnitude comparisons using L2 Norm}
\begin{figure*}[!ht]
    \centering
    \includegraphics[width=\textwidth]{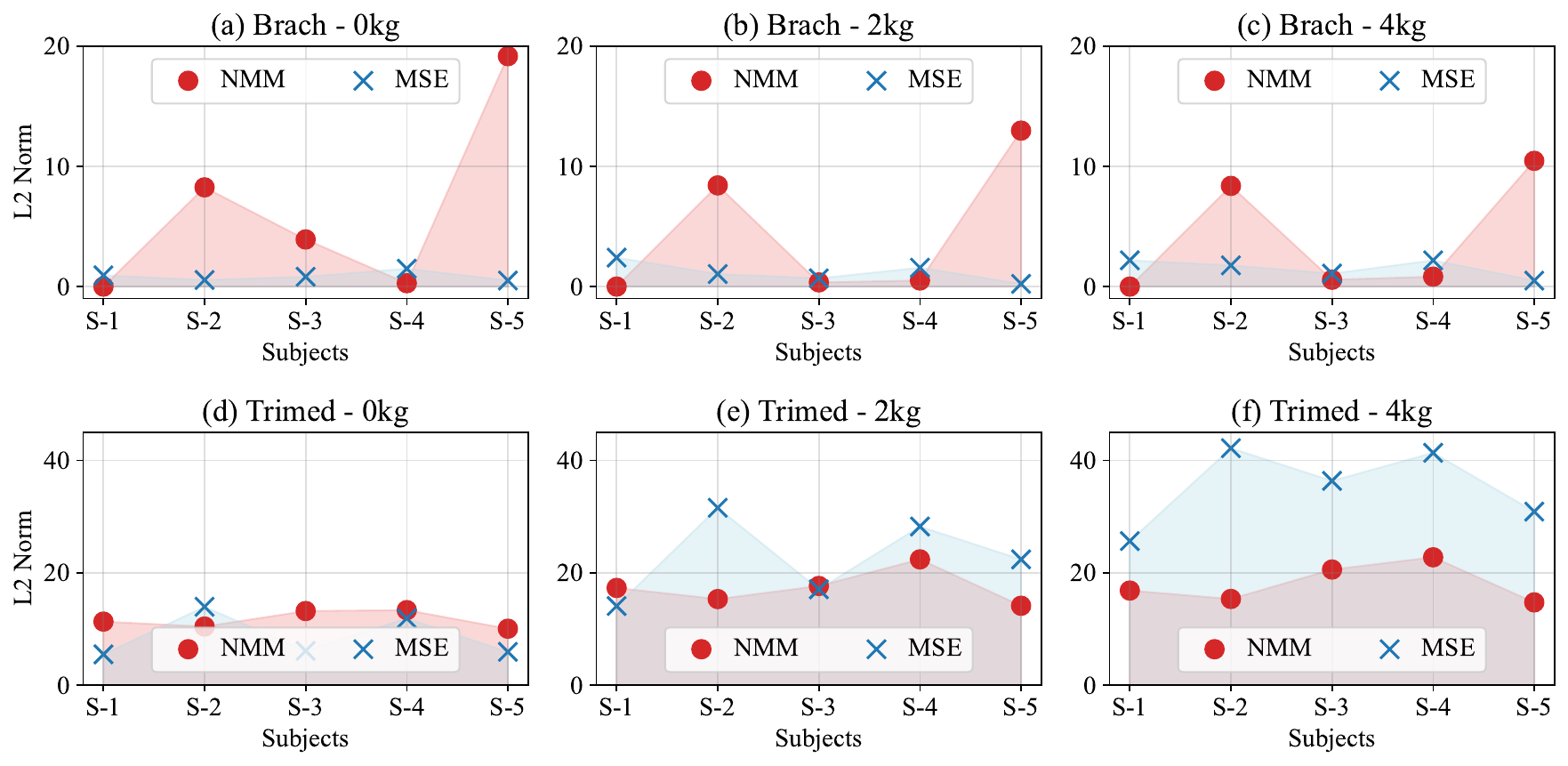}
    \caption{L2 norm comparisons between NMM and MSE for the (a, b, and c) Brachialis (Brach) and (d, e, and f) Triceps medial head (Tri med) muscles across different subjects and load conditions. The scatter plots display the L2 norms for both methods, with red circles representing NMM and blue crosses representing MSE, highlighting the differences in their magnitude for EMG signal prediction at varying loads.}
    \label{fig:L2norm}
\end{figure*}

The L2 norm was computed for both the constructed EMG signal envelopes, providing a quantitative measure of the overall signal strength.
These values are illustrated in Fig. \ref{fig:L2norm} as a scatter plot for each subject, muscle, and load condition. Each row represents a specific muscle on different loads, with red circles and blue crosses denoting the methods NMM and MSE, respectively. The y-axis displays the L2 norm values, while the x-axis represents the subjects (S-1 to S-5).
Using the proposed NMM, the EMG amplitude for the \textit{Brach} muscle is nearly zero for subjects S-1, S-3, and S-4, while subjects S-2 and S-5 show slightly higher values. For the \textit{Trimed} muscle, the EMG amplitude is consistent across all subjects (S-1 to S-5) and increases with load, reflecting physiological trends. The consistency in the EMG amplitude ensures that the constructed envelopes are within physiologically plausible ranges and reflect realistic muscle activation levels.
A comparison of L2 norm values with the MSE (blue crosses) is also shown in Fig. \ref{fig:L2norm}, highlighting notable differences between the two methods. Furthermore, a paired t-test (\(p < 0.001\)) confirms significant differences in prediction magnitudes between NMM and MSE. In particular, NMM demonstrates a lower EMG amplitude for the \textit{Trimed} muscle at higher loads compared to MSE. These deviations from the MSE results highlight areas where the proposed method diverges in approximating muscle behavior.
\subsection {Joint torque validation}
\begin{figure*}[!ht]
    \centering
    \includegraphics[width=\textwidth]{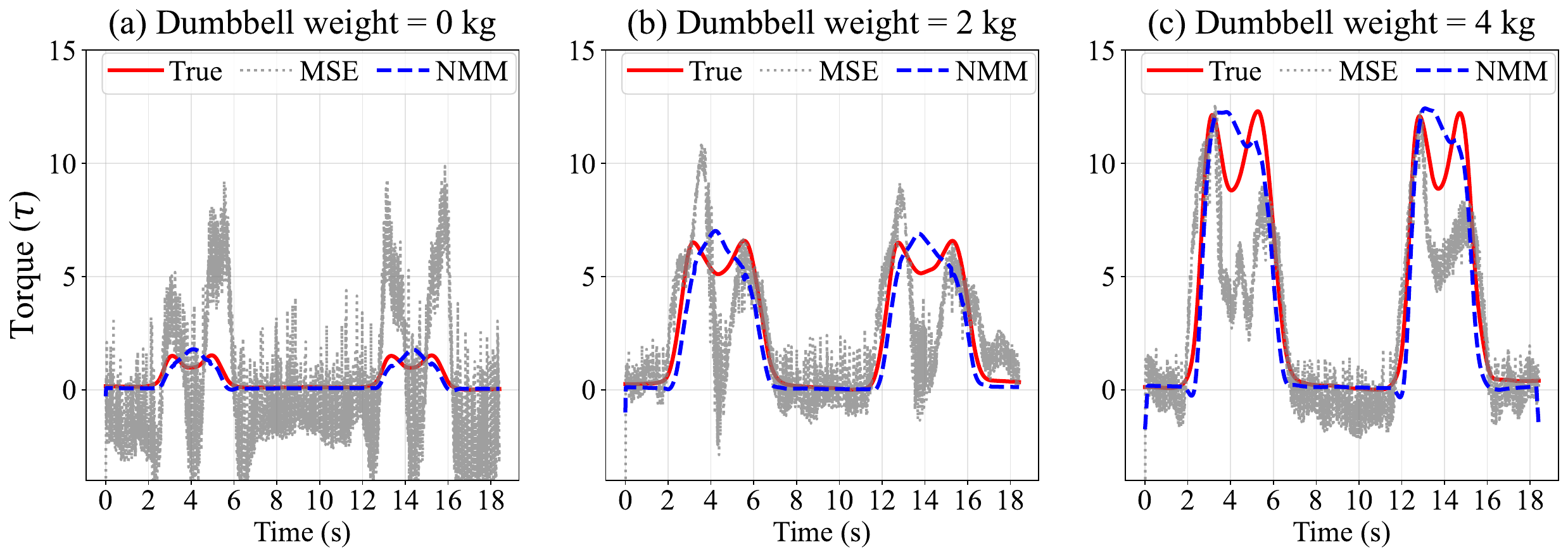}
    \caption{Comparison of the torque profiles for the measured and constructed EMG signals for a representative test subject. The torque profiles are predicted in real-time for different loads from the EMG predictions of the proposed neural musculoskeletal model.}
    \label{fig:torque_pinn}
\end{figure*}

\begin{figure*}[!ht]
    \centering
    \includegraphics[width=\textwidth]{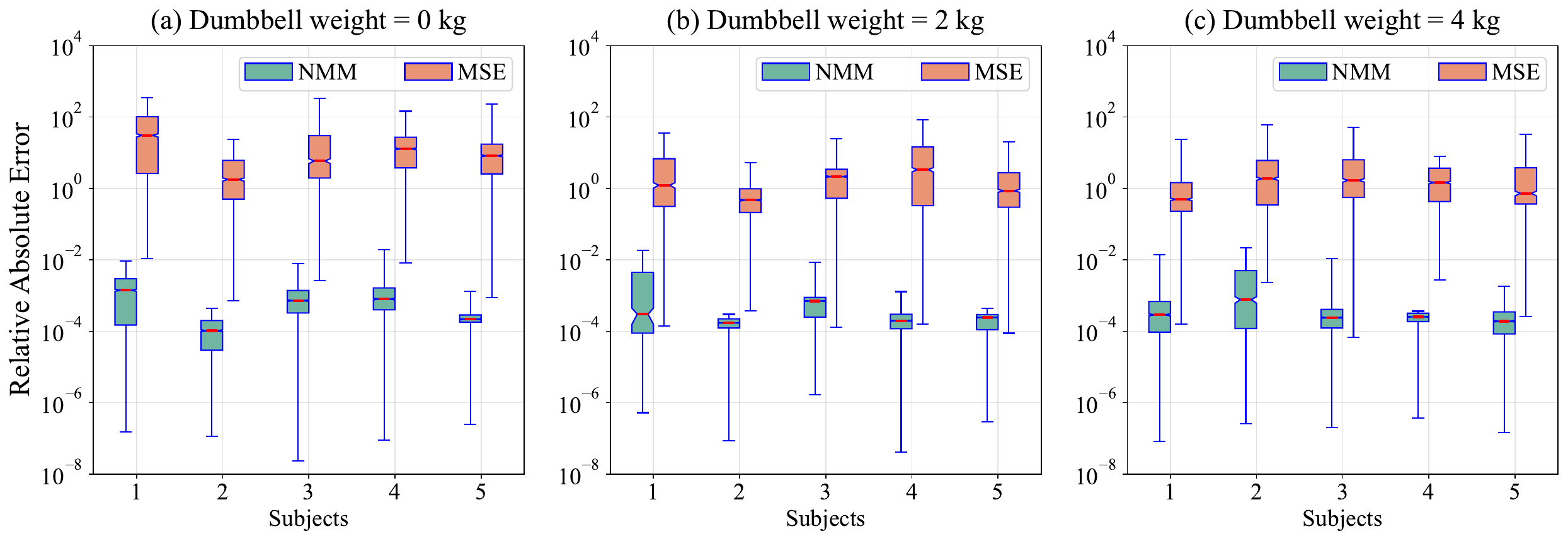}
    \caption{Comparison of the torque profiles for the measured and constructed EMG signals. The torque profiles are predicted in real-time for different loads from the EMG predictions of the proposed neural musculoskeletal model.}
    \label{fig:error_torque}
\end{figure*}

\begin{figure*}[!ht]
    \centering
    \includegraphics[width=\textwidth]{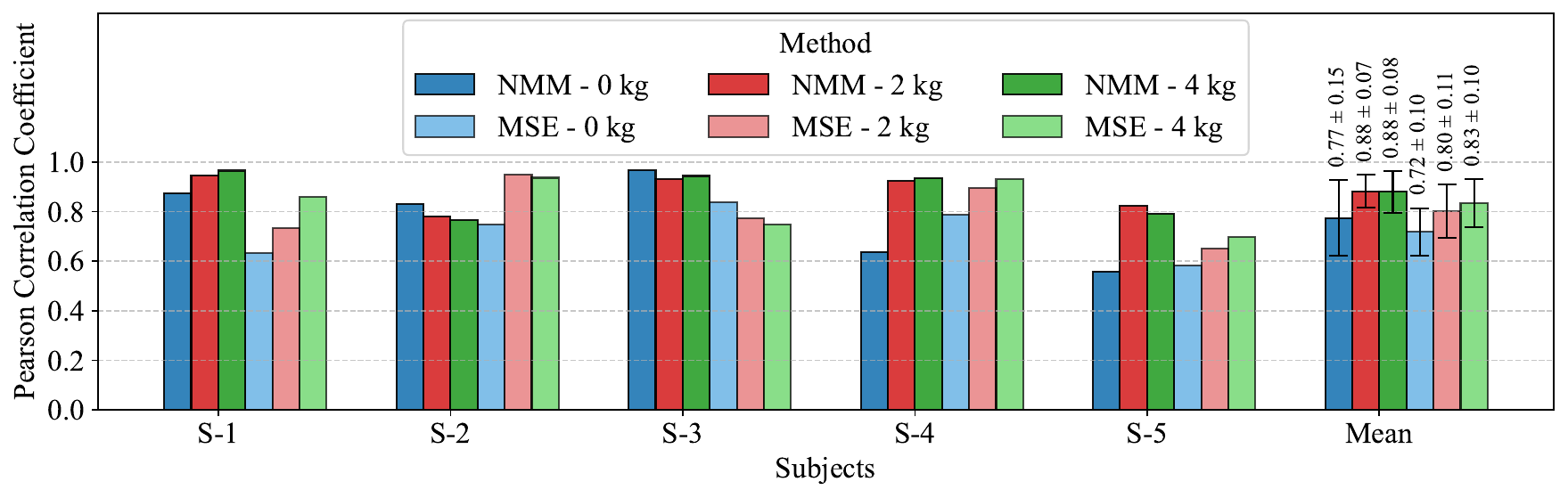}
    \caption{Comparison of the torque profiles for the measured and constructed EMG signals. The torque profiles are predicted in real-time for different loads from the EMG predictions of the proposed neural musculoskeletal model.}
    \label{fig:PCCR}
\end{figure*}

To evaluate the performance of the NMM in constructing the \textit{Brach} and \textit{Trimed} muscle EMG signal envelopes and its physiological precision, we further estimated the torque of the elbow joint. We combined the constructed EMG envelope with the measured ones for all load conditions and used it as input to the personalized MSK model to calculate joint torque.
Fig. \ref{fig:torque_pinn} illustrates the torque profiles for a representative test subject, showing the fit under different load conditions (0 kg, 2 kg, and 4 kg) using both the NMM and MSE methods, with the true torque values derived from the inverse dynamics tool in `OpenSim' software \cite{delp2007opensim}. To compute the corresponding ground truth joint torque, we added a dumbbell with different loads to the arm of the scaled 'arm26.osim' model in `OpenSim' software, which takes the joint angle as input. 
To evaluate the accuracy and robustness of the model across subjects and loads, we calculated the RAE and PCC between the NMM-estimated joint torque and the true inverse dynamics values, as shown in Figs. \ref{fig:error_torque} and \ref{fig:PCCR}. 
For the NMM, the average PCC values for elbow joint torque across five participants were 0.77$\pm$0.15, 0.88$\pm$0.07, and 0.88$\pm$0.08 for 0, 2, and 4 kg loads, respectively. Similarly, the average RAE values across the participants were 0.02$\pm$0.81, 0.01$\pm$0.55, and 0.02$\pm$0.55 for the same load conditions. 
In comparison, for the MSE method, the average PCC values were 0.72$\pm$0.10, 0.80$\pm$0.11, and 0.83$\pm$0.10 for 0, 2, and 4 kg loads, respectively, with average RAE values of 883$\pm 38 \times 10^{3}$, 76$\pm 2.8 \times 10^{3}$, and 62$\pm 1.4 \times 10^{3}$. These results highlight the robustness of the NMM in predicting EMG signals and highlight the functional significance of the constructed EMG in movement analysis.
\subsection {Physiological MSK parameter estimation}
\begin{figure*}[!ht]
    \centering
    \includegraphics[width=\textwidth]{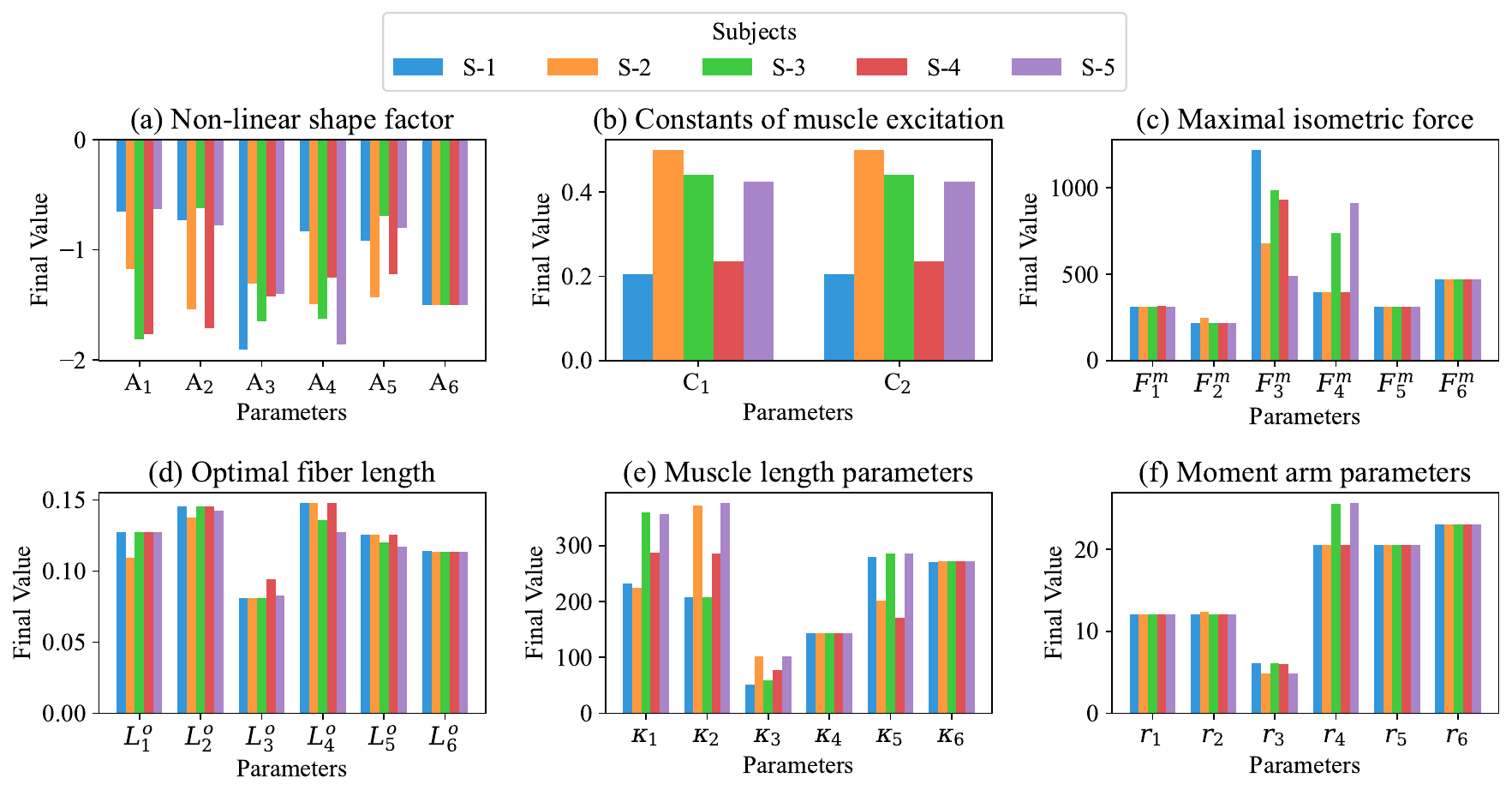}
    \caption{Final stationary values of the physical parameters of the neural musculoskeletal model for all five subjects. The parameters are optimized for all six muscles.}
    \label{fig:finalparameters}
\end{figure*}

\begin{figure*}[!ht]
    \centering
    \includegraphics[width=\textwidth]{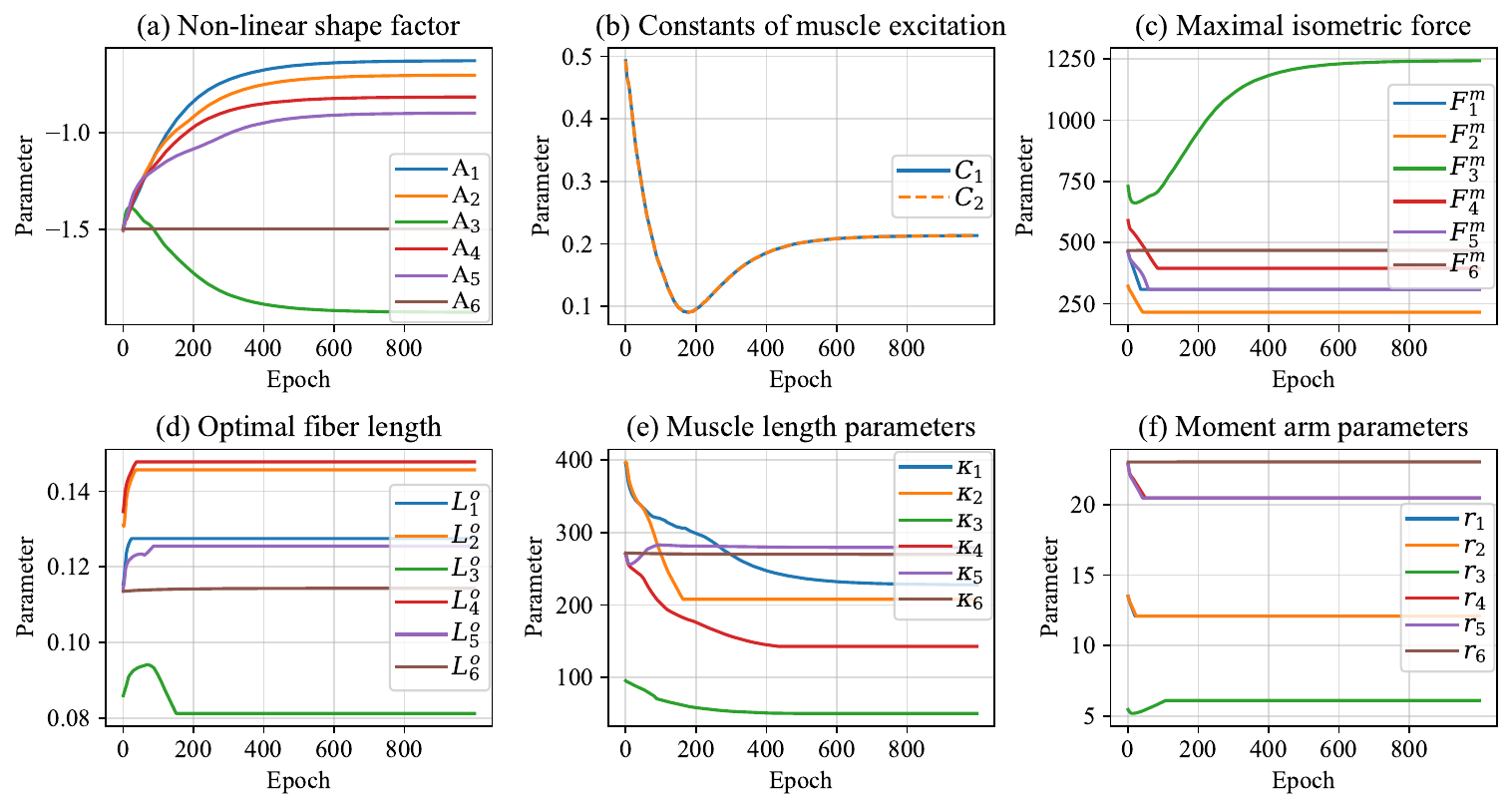}
    \caption{Evolution of the physical parameters of the neural musculoskeletal model. The parameters are optimized for all six muscles. All the muscle parameters start approaching the stationary values simultaneously.}
    \label{fig:parameters}
\end{figure*}

During NMM training, we also optimized 32 parameters of the MSK model (across six units of muscles) for each subject to allow the model to accurately construct the EMG data for the deep muscles. Fig. \ref{fig:finalparameters} illustrates the final subject-specific physiological parameters determined for the five subjects (S-1 to S-5) through the training process. The evolution of these estimated parameters during training is shown in Fig. \ref{fig:parameters} for a representative subject. Detailed information on the parameters and their optimized values for the representative subject is provided in Table \ref{tab:MSKparameters}. 
The physiological ranges of the MSK parameters were derived from \cite{holzbaur2005model,lloyd2003emg,pigeon1996moment}. As seen in Table \ref{tab:MSKparameters} and Fig. \ref{fig:finalparameters}, the identified physiological parameters remain within the acceptable physiological range and are consistent across subjects. This confirms that the personalized MSK model integrated into the proposed method produces reasonable constructions of muscle EMGs.
\begin{table}[h!]
\centering
\caption{Estimated physiological parameters of the MSK model for the representative subject.}
\resizebox{\columnwidth}{!}{%
\begin{tabular}{c c c c c }
\hline
\textbf{MSK Parameter} & \textbf{Description}                                & \textbf{Muscle Name} & \textbf{Range} & \textbf{Optimized Value} \\ \hline
\multirow{6}{*}{\( F_o^m \) (N)} & \multirow{6}{*}{\centering Max. isometric force} & Biclong           & 312.17--936.45 & 312.17 \\  
                              &                                                 & Bicshort          & 217.80--653.40  & 217.80 \\ 
                              &                                                 & Brach       & 493.65--1480.95   & 1215.85 \\ 
                              &                                                 & Trilong  & 399.25--1197.75   & 395.25 \\ 
                              &                                                 & Trilat          & 312.17--936.45 & 312.17 \\ 
                              &                                                 & Trimed         & 312.17--936.45  & 468.22 \\ \hline
                              
\multirow{6}{*}{\( L_m^o \)(m)} & \multirow{6}{*}{\centering Optimal fiber length}    & Biclong           & 0.1131--0.1246 & 0.1246 \\  
                              &                                                 & Bicshort          & 0.1287--0.1420 & 0.1420 \\ 
                              &                                                 & Brachialis       & 0.0838--0.0882 & 0.079 \\ 
                              &                                                 & Trilong  & 0.1306--0.1441 & 0.1441 \\ 
                              &                                                 & Trilat          & 0.1112--0.1226 & 0.1226 \\ 
                              &                                                 & Trimed         & 0.1112--0.1168 & 0.1116 \\ \hline
                              
\multirow{6}{*}{\( \text{A} \)} & \multirow{6}{*}{\centering Non-linear activation shape factor} & Biclong     & -3.0--0.0 & -0.654 \\
                              &                                                 & Bicshort          & -3.0--0.0 & -0.727 \\ 
                              &                                                 & Brachialis       & -3.0--0.0 & -1.910 \\ 
                              &                                                 & Trilong  & -3.0--0.0 & -0.832 \\ 
                              &                                                 & Trilat          & -3.0--0.0 & -0.919 \\ 
                              &                                                 & Trimed         & -3.0--0.0 & -1.500 \\ \hline

\( \text{$C_1$, $C_2$} \)    & \centering Constants of excitation          & All muscles     &  -1.0--1.0 & 0.204 \\ \hline

\multirow{6}{*}{\( \kappa \) (cm)} & \multirow{6}{*}{\centering muscle Length parameter} & Biclong           & 207.93--415.86   & 231.21 \\  
                              &                                                 & Bicshort          & 207.93--415.86   & 207.93 \\ 
                              &                                                 & Brach       & 50.61--101.22   & 50.61 \\ 
                              &                                                 & Trilong  & 143.02--286.05   & 143.02 \\ 
                              &                                                 & Trilat          & 143.02--286.05   & 279.18 \\ 
                              &                                                 & Trimed         & 143.02--286.05   & 269.74 \\ \hline
                              
\multirow{6}{*}{\( \text{r} \) (cm)} & \multirow{6}{*}{\centering Moment arm parameter}  & Biclong           & 12.09--15.12 & 12.09 \\  
                                &                                               & Bicshort          & 12.09--15.12 & 12.09 \\ 
                                &                                               & Brach       & 4.87--6.09 & 6.09 \\ 
                                &                                               & Trilong  & 20.49--25.61 & 20.49 \\ 
                                &                                               & Trilat          & 20.49--25.61 & 20.49 \\ 
                                &                                               & Trimed         & 20.49--25.61 & 23.05 \\ \hline
\end{tabular}%
}
\label{tab:MSKparameters}
\end{table}
\section{Discussion}\label{sec:discussion}
This study presents a novel physics-integrated neural network approach, termed NMM, for constructing unmeasured EMG signal envelopes of deep muscles. The proposed NMM uses joint kinematic states and lifted load information as inputs, encoding essential movement features and personalized musculoskeletal (MSK) model characteristics to generate accurate EMG predictions. To the best of our knowledge, this is the first application of an MSK physics-integrated inverse simulation for constructing unmeasured deep muscle EMG signals. 

The NMM successfully constructed EMG estimates for upper limb deep muscles, specifically the \textit{Brachialis} and \textit{Triceps medial head}, using elbow joint angles and angular velocity data recorded during elbow flexion-extension movements. The NMM demonstrated robust performance across dynamic movements with three different loading conditions (0, 2, and 4 kg), outperforming traditional matrix factorization-based MSE approaches in accuracy.
The analysis of the results for the five subjects clearly showed that NMM is able to generate estimates of unmeasured muscle EMG envelopes that are consistent in spatio-temporal patterns and magnitudes of muscle activation. The heatmap Fig. \ref{fig:Heatmap} and L2 norm Fig. \ref{fig:L2norm} comparison of constructed EMGs clearly demonstrate that the performance of the NMM surpasses the state-of-the-art MSE approach, particularly in predicting the EMG of the \textit{triceps medial head} muscle at higher loads. The EMG envelope predicted by the NMM remained within a maximum value of 1, whereas the MSE approach did not exhibit this consistency.

The results show that the L2 norm value of the EMG envelope of constructed muscle brachialis (\textit{Brach}) is close to zero (see Fig. \ref{fig:L2norm}) for all the loading conditions. This means \textit{Brach} muscle is not being recruited in the eFE movement experiment considered in this study. This aligns with the results of experimental research on the major muscles contributing to the elbow flexion motion \cite{palazzi2006neurotization}. On the other hand, the results of NMM suggest that a significant triceps medial head (\textit{Trimed}) muscle activation is found, with activation level increasing as the load increases. However, this needs an experimental verification.
Furthermore, the comparison of joint torque (Figs. \ref{fig:torque_pinn}, \ref{fig:error_torque}, \ref{fig:PCCR}) obtained from constructed EMGs shows that the NMM provides more dynamically consistent and accurate joint torque compared to the MSE approach. Highlighting the functional significance of the NMM-constructed EMG envelope in movement analysis.

Unlike MSE, which decomposes measurable EMG data from four surface muscles to derive basis synergy vectors and weights, neglecting deep muscle EMGs and constructing missing muscle signals based on incomplete data. The proposed method constructs EMG entirely from the joint kinematic state and lifted load. Additionally, because the basis used in MSE lacks comprehensive information, the constructed EMGs for missing muscles are less accurate. In contrast, the proposed NMM uses the measurable joint kinematic state (joint angle and angular velocity) to provide a more reliable solution for constructing deep or missing muscle EMGs.

In this study, angular velocity was derived by calculating the first-order derivative of joint angles with respect to time, obtained from motion capture data. While this approach is effective for the current analysis, it relies heavily on the precision of numerical differentiation, which may introduce noise and affect the accuracy of the angular velocity estimation.  To enhance the reliability and efficiency of data collection in future studies, we propose the integration of inertial measurement units (IMUs) for direct measurement of both joint angles and angular velocities \cite{seel2014imu}. IMUs offer the advantage of real-time data acquisition and eliminate the need for numerical differentiation, potentially improving the accuracy of dynamic motion analysis \cite{gu2023imu}. This advancement would also allow for broader applications of NMM in settings where motion capture systems are impractical or unavailable, enabling portable and real-time application of the proposed approach. In addition, since the unmeasurable EMG signal and the corresponding torques are predicted from joint kinematics and load information. The proposed NMM also provides a possible adoption for predicting EMG envelopes of missing muscle or lower limb deep during gait analysis. During gait motion, joint kinematics and ground reaction force measurement are very common, which can be directly used as input to the NMM.

In this study, we used an attention-based transformer neural network integrated with forward musculoskeletal dynamics to construct deep muscle EMG patterns while simultaneously optimizing subject-specific physiological parameters. The transformer was chosen for its exceptional ability to learn long-range dependencies between EMG, joint kinematics, and dynamic loading conditions  through the self-attention mechanism, which is critical to capturing the intricate temporal and spatial relationships inherent in joint kinematics, EMG data, and the forward dynamic model. 
However, the current NMM architecture's training time is a bit longer, considering only two deep muscles, taking approximately 4 hours to complete 1000 epochs on a 12-core CPU. This could be due to the fact that the computation of the physics loss $\mathcal{L}_{\mathrm{phy.}}$ in Eq. \eqref{eq:combined_loss} involves evaluation of the forward MSK model, in particular the Eqns. \eqref{Eqn: force velocity}, \eqref{Eqn: passive force len}.
These equations are modeled as a piecewise function with each segment of the function describing the behavior under different conditions (e.g., shortening, lengthening, or isometric phases). Piecewise definitions introduce sharp transitions, making it harder for neural networks to approximate the function smoothly. Additionally, gradients become more complex to compute, especially at boundaries, complicating optimization. 
Strategies like using spline-based approximations to the piecewise function can mitigate these challenges to some degree. Nevertheless, further study is required to address the bottleneck of the larger training time.

In implementing the proposed method, we personalized a generic musculoskeletal (MSK) forward dynamics model while training the physics-integrated neural network. The optimized personalized physiological parameters for six muscle units are provided and described in Table \eqref{tab:MSKparameters} for a representative subject. In the study, we have used six primary upper arm muscles as the main actuators for elbow flexion and extension. These muscles can also influence other degrees of freedom (DOF) of shoulder movement, which has been neglected. Similarly, the lower arm muscles may also influence eFE movement, which has been neglected here. Future studies can also include shoulder DOF and muscles from the lower arm. This would enable a more accurate representation of MSK system dynamics, enhancing its accuracy and practicality for clinical applications.

\section{Conclusion} \label{conclusion}\label{sec:conclusion}
In summary, the proposed NMM presents an innovative and non-invasive approach for constructing deep or missing muscle EMG signals using only joint kinematics data and lifted load. Our findings show that NMM effectively captures the spatial distribution and temporal patterns of EMG envelopes, highlighting its capability to encode essential features across both the time and frequency domains. This is achieved by training an attention-based Transformer neural network integrated with a personalized musculoskeletal (MSK) forward dynamics model.

The proposed NMM framework has been validated on self-collected datasets from five test subjects performing upper limb tasks and compared against the MSE approach. The NMM architecture effectively constructed the EMG envelopes of the \textit{Brachialis} and \textit{Triceps medial head} muscles using elbow joint kinematics across various loading conditions, outperforming the state-of-the-art method. However, further study is required to improve the accuracy and generalizability of NMM across diverse populations in generating signals from multiple muscle groups simultaneously across multiple joints with diverse movement patterns in both upper and lower limbs. To this end, it may require to design more comprehensive experimental protocols to enhance data quality and incorporate a more precise MSK model—especially refining the muscle contraction dynamics and geometry modules within the physics-encoding block. 

The NMM only requires joint kinematic measurements and load data after training in order to predict the EMG envelope. Joint kinematics and load data collection are frequently used in offline and online movement analysis, with the possibility of making the proposed approach portable and real-time. Due to this benefit, it could be a useful alternative to invasive methods and be applied in rehabilitation engineering and clinical settings. Furthermore, the proposed method might be extended to generate raw EMG signals and other neuromuscular signals, like motor unit action potentials, which could offer a more comprehensive understanding of neuromuscular dynamics and motor control.
\section*{Acknowledgment} \label{Acknowledgment}
The authors would like to thank Anant Jain from Indian Institute of Technology (IIT) Delhi for helping in data collection. This study is partly supported by the Joint Advanced Technology Centre (JATC) (Grant no.: RP03830G) and the I-Hub Foundation for cobotics (IHFC) section-8 company (Grant no.: GP/2021/RR/010) at IIT Delhi, sponsored by the Ministry of Education (MoE), Govt. of India.

\clearpage

\appendix

\section*{Appendix}

\section{Hyperparameter details of Transformer}
There are four encoder $\mathcal{E}_j$ and decoder $\mathrm{D}_j$ blocks in the transformer architecture. For consistency in the embedding dimension, the configuration of the encoder and decoder transformer blocks is kept the same. The hyperparameters are provided in Table \ref{tab:tarnsformer}.
For optimization of the neural parameters $\bm{\theta}$ and the additional 32 parameters, we use the AdamW optimizer with the configuration provided in Table \ref{tab:optimizer}.
\begin{table}[!ht]
    \centering
    \caption{Encoder and Decoder Transformer Configuration}
    \label{tab:tarnsformer}
    \begin{tabular}{lc}
        \toprule
        Attention parameters & Value \\
        \midrule
        Heads in Multi-Head Attention & 12 \\
        Input/Output dimension of each layer & 4 \\
        Dimension of Query/Key/Value & 32 \\
        Feed-Forward layers & 5 \\
        Hidden layer dimension & [64,128,256,128,64] \\
        \bottomrule
    \end{tabular}
\end{table}
\begin{table}[!ht]
    \centering
    \caption{Configuration of Optimizer and Learning Rate Schedule}
    \label{tab:optimizer}
    \begin{tabular}{lc}
        \toprule
        Optimizer parameters & Value \\
        \midrule
        Initial learning rate & $10^{-4}$ \\
        Learning rate scheduler & Step scheduler \\
        Scheduler step & 25 \\
        Scheduler factor & 0.8 \\
        Weight decay & $10^{-4}$ \\
        AdamW momentum parameters ($\beta_1$, $\beta_2$) & 0.9, 0.99 \\
        Epoch size & 500 \\
        Batch size & 100 \\
        \bottomrule
    \end{tabular}
\end{table}

\section{Algorithm and Hyperparameters of the MSE approach}
The pseudo-code for implementing the MSE framework is briefly illustrated in the Algorithm \ref{algo:MSE}). For consistency in comparison, we optimize the MSE parameters: \( \mathbf{S} \in \mathbb{R}^{\text{len\_data} \times N_s} \) and \( \mathbf{W} \in \mathbb{R}^{N_s \times N_m} \) using the Adam optimizer with a mean squared error (MSE) loss function. 
The optimization parameters are provided in Table \ref{tab:MASEparameters}.
\begin{algorithm}[!t]
    \caption{Muscle Synergy Extrapolation with Joint Torque Minimization}
    \label{algo:MSE}
    \begin{algorithmic}[1]
    \State \textbf{Input:} Measured EMG matrix \( \mathbf{M_m} \), inverse dynamics torque \( \mathbf{\tau_{\text{ID}}} \), optimized MSK parameters.
    \State \textbf{Output:} constructed EMG matrix \( \mathbf{A} = \text{stack}(\mathbf{M_m}, \mathbf{E_r}) \)
    
    \State Set number of synergies \( N_s \)
    \State Initialize \( \mathbf{S_m}, \mathbf{W_m}, \mathbf{W_r} \) (random initialization, enforce non-negativity)
    \State Initialize Adam optimizer for \( \mathbf{S_m}, \mathbf{W_m}, \mathbf{W_r} \)
    
    \For{iterations \( \textit{NumIters1} \)}
        \State Perform Non-Negative Matrix Factorization (NMF) on \( \mathbf{M_m} \):
        \State \quad Extract synergies \( \mathbf{S_m} \) and coefficients \( \mathbf{W_m} \) such that:
        \State \quad \( \mathbf{M_m} \approx \mathbf{S_m} \cdot \mathbf{W_m} \)
        \State Compute construction error \( \mathbf{\epsilon_m} = \frac{1}{n} \sum || \mathbf{M_m} - \mathbf{S_m} \cdot \mathbf{W_m} ||^2 \) \Comment{Mean squared error}
        
        \If{\( \mathbf{\epsilon_m} \leq \textit{tol\_m} \)} \Comment{Stopping criterion for NMF}
            \State \textbf{Break}
        \EndIf
        
        \For{iterations \( \textit{NumIters2} \)}
            \State construction unmeasured EMG \( \mathbf{E_r} \):
            \State \quad \( \mathbf{E_r} \approx \mathbf{S_m} \cdot \mathbf{W_r} \)
            \State matrix \( \mathbf{A} = \text{stack}(\mathbf{M_m}, \mathbf{E_r}) \) \Comment{Stack measured and constructioned EMG}
            
            \State Compute predicted torque \( \mathbf{\tau_{\text{pred}}} \) using \( \mathbf{A} \) and MSK parameters.
            \State Compute torque error \( \mathbf{\epsilon_{\tau}} = \frac{1}{n} \sum || \mathbf{\tau_{\text{pred}}} - \mathbf{\tau_{\text{ID}}} ||^2 \) \Comment{Mean squared error}
            
            \If{\( \mathbf{\epsilon_{\tau}} \leq \textit{tol\_tau} \)} \Comment{Stopping criterion for torque minimization}
                \State \textbf{Break}
            \EndIf
        \EndFor
        
        \State Update \( \mathbf{S_m}, \mathbf{W_m}, \mathbf{W_r} \) to jointly minimize \( \mathbf{\epsilon_m} \) and \( \mathbf{\epsilon_{\tau}} \) using the Adam optimizer.
    \EndFor
    \end{algorithmic}
\end{algorithm}
\begin{table}[ht!]
    \centering
    \caption{Parameters for the MSE Algorithm}
    \begin{tabular}{c c l}
    \hline
    \text{Parameter} & \text{Value} & \text{Description} \\ \hline
    \( N_m \) & 4 & Number of muscle EMG measured. \\
    \( N_s \) & 3 & Number of muscle synergies. \\
    \( \textit{NumIters1} \) & 4000 & Max. number of iterations to perform NMF. \\
    \( \textit{NumIters2} \) & 2000 & Max. number of iterations for torque minimization. \\
    \( \text{Optimizer} \) & Adam & Optimization algorithm used. \\
    \( \text{learning\_rate} \) & $10^{-3}$ & Learning rate for the optimizer. \\
    \( \textit{tol\_m} \) & $10^{-6}$ & Stopping criteria for NMF. \\
    \( \textit{tol\_tau} \) & $10^{-1}$ & Stopping criteria for torque minimization. \\
    \( \text{len\_data} \) & Length of test dataset & Total size of the dataset being processed. \\
    \hline
    \end{tabular}%
    \label{tab:MASEparameters}
\end{table}


\end{document}